%% file: main.tex
\documentclass[runningheads]{llncs}

\input{preamble}
\begin{document}

\title{StrucTab: A Structured Optimization Framework for Table Parsing}

\input{section/_1_author}

\maketitle

\input{section/_1_author_footnote}
\input{section/0_abstract}
\input{section/1_introduction}
\input{section/2_background}
\input{section/3_StrucTab}
\input{section/4_TableVerse}
\input{section/5_experiments}
\input{section/6_conclusion}

\section*{Acknowledgements}

Supported by the National Natural Science Foundation of China (Grant NO 62376266 and 62406318) and CAAI-Tencent Rhino-Bird Open Research Fund.

\newpage
\bibliographystyle{splncs04}
\bibliography{main}

\input{section/7_appendix}

% \newpage
% \bibliographystyle{splncs04}
% \bibliography{main}

\end{document}

%% file: preamble.tex
\usepackage{eccv}

\usepackage{eccvabbrv}

\usepackage{graphicx}
\usepackage{booktabs}
\usepackage[accsupp]{axessibility}  % Improves PDF readability for those with disabilities.

\usepackage{hyperref}

\usepackage{orcidlink}
\usepackage{multirow}
\usepackage{makecell}
\usepackage{amssymb}
\usepackage{booktabs}
\usepackage{graphicx}
\usepackage{caption}
\usepackage{pifont}
\usepackage{enumitem}
\usepackage{amsmath}
\usepackage{algorithm}
\usepackage{algorithmicx}
\usepackage{algpseudocode}
\usepackage{fontawesome5}

\newcommand{\cmark}{\ding{51}}
\newcommand{\xmark}{\ding{55}}

\newcommand{\mypara}[1]{\smallskip\noindent\textbf{#1}\hspace{0.02cm}}

\let\titleold\title
\renewcommand{\title}[1]{\titleold{#1}\newcommand{\thetitle}{#1}}
\def\maketitlesupplementary
{
    {
            \newpage
            \par\nolinenumbers
            \centering
            \Large
            \textbf{\thetitle}\\
            \vspace{0.3em}
            Supplementary Material \\
            \linenumbers
        }
}

\newcommand{\nextline}{\\}

\definecolor{table_ours}{HTML}{F5FFFA}   % row highlighting in result tables

%% file: section/_1_author.tex
\author{
    Gengluo Li\inst{1,4}$^{,\scalebox{1}{$\star$}}$
    % Gengluo Li\inst{1,4}$^{,\scalebox{1}}$\thanks{Equal contribution.}
    \and
    Shangpin Peng\inst{2,5,6}$^{,\scalebox{1}{$\star$}}$
    \and
    Chengquan Zhang\inst{2}$^{,\dagger}$
    \and
    Binghong Wu\inst{2}
    \and
    \\[0.2em]
    Hao Feng\inst{2}
    \and
    Weinong Wang\inst{2}
    \and
    Pengyuan Lyu\inst{2}
    \and
    Huawen Shen\inst{1,4}
    \and
    Xingyu Wan\inst{2}
    \and
    \\[0.2em]
    Zhuotao Tian\inst{5}
    \and
    Han Hu\inst{2}
    \and
    Can Ma\inst{1,4}
    \and
    Yu Zhou\inst{3,\,}\textsuperscript{\scalebox{0.8}{\faEnvelope}}
}

\authorrunning{G.~Li et al.}

\institute{
    Institute of Information Engineering, Chinese Academy of Sciences, Beijing, China
    \and
    LLM Department, Tencent, China
    \and
    Nankai University, China
    \and
    School of Cyber Security, University of Chinese Academy of Sciences, China
    \and
    Shenzhen Loop Area Institute, China
    \and
    Hong Kong University of Science and Technology, China
    \\[0.4em]
    \email{
        ligengluo@iie.ac.cn
        \quad
        pspdada0808@gmail.com
        \quad
        yzhou@nankai.edu.cn
    }
}

%% file: section/_1_author_footnote.tex
\renewcommand{\thefootnote}{}
\footnotetext{\footnotesize
    $^{\scalebox{1}{$\star$}}$~Equal contribution.
    \quad
    $^{\dagger}$~Project leader.
    \quad
    \textsuperscript{\scalebox{1}{\faEnvelope}~}Corresponding author.
}
\renewcommand{\thefootnote}{\arabic{footnote}}

%% file: section/0_abstract.tex
\begin{abstract}
    Table parsing aims to convert table images into structured, machine-readable representations, a task requiring the joint perception of complex spatial layouts and textual content. While recent vision-language models (VLMs) enable end-to-end parsing, they typically rely on direct supervision of the final output, thereby bypassing the explicit intermediate reasoning that is crucial for understanding complex table structures. Furthermore, attempts to optimize these models using reinforcement learning (RL) are often hindered by unstable or ambiguous reward designs, limiting potential performance gains.
    To address these limitations, we propose \textbf{StrucTab}, a table parsing model learned through intermediate structural supervision and reward decomposition.
    At the modeling level, by decomposing the parsing process into human-inspired subtasks, such as row--column counting and merged-cell analysis, StrucTab progressively unifies them through a sequential reasoning strategy.
    At the optimization level, we introduce \textbf{Uni-TabRL}, a unified RL framework that leverages decomposed rewards (validity, structure, and content) to provide stable and informative optimization signals.
    Finally, at the evaluation level, we present \textbf{TableVerse-5K}, a large-scale, challenging benchmark encompassing diverse, real-world table scenarios.
    Extensive experiments demonstrate the state-of-the-art performance of StrucTab across all evaluated public benchmarks and significant improvements on TableVerse-5K, validating the effectiveness of explicit structural modeling and decomposed reward optimization.
    Code and benchmark are publicly available at \url{https://github.com/VirtualLUOUCAS/StrucTab}.
    \keywords{Table Parsing \and VLMs \and Structured Reasoning \and RL}
\end{abstract}

%% file: section/1_introduction.tex
\section{Introduction}
\label{sec:intro}

Table parsing is a fundamental problem in document understanding, which aims to convert table images into structured, machine-readable representations such as HTML or Markdown~\cite{shen2023divide, NCGM_2022, TDBU_2020, LORE_2023, Document_parsing_2024}. Unlike plain text, tables exhibit complex spatial layouts and logical relationships, requiring models to jointly perceive both structural organization and textual content~\cite{TabPedia_2024, Image_table_2020, TFLOP_2025, Infinity_parser_2025}. This intrinsic complexity makes robust table parsing particularly challenging in real-world scenarios.

\input{figure_code/introduction}

The methodology for table parsing has undergone a significant paradigm shift from cascaded pipelines to end-to-end generation. Traditionally, this task relied on cascaded pipelines that decoupled structure recognition from optical character recognition (OCR)~\cite{PP_OCRv2_2021, 2025_pacm, Nougat_2024, Vary_2024, Mineru_2024, jiahao_2025}. While effective in constrained settings, such cascaded designs inevitably suffer from error propagation and limited robustness in diverse, in-the-wild layouts~\cite{parsing_table_wild_2021}. To address these limitations, VLMs have revolutionized the field by directly parsing images into structured codes~\cite{Real_World_Doc_2026, LightOnOCR_2026, Dots_ocr_2025, Logics_parsing_2025, olmOCR_2025}. However, as shown in~\cref{fig:introduction} (a), although this unified modeling simplifies the pipeline, it primarily relies on the autoregressive generation of a linearized sequence, treating table parsing as a flat, one-dimensional image-to-text task. Since tables are inherently structured entities where logical relationships among rows, columns, and merged cells are paramount, this reliance on end-to-end generation raises a fundamental question: \textit{Can we truly expect a model to internalize the rigorous topological logic of complex tables solely by mimicking the final serialized output, without explicitly modeling the intermediate reasoning process?}

To answer this, we revisit task modeling through a cognitive lens. From a human perspective, table parsing is rarely a single-step inference. Instead, as illustrated in~\cref{fig:introduction} (b), it follows an explicit reasoning trajectory: first perceiving the global layout, then resolving local dependencies such as merged cells, and finally reconstructing the content. Conversely, current VLM-based methods \textit{bypass these intermediate steps}, thereby forcing the model to implicitly infer complex structural logic from the final output. This black-box approach lacks structural scaffolding, making optimization inefficient and leading to suboptimal generalization.

\input{figure_code/RL_related_work}

Beyond task modeling, effectively optimizing table parsing models requires advanced learning paradigms. Reinforcement learning (RL) has emerged as a promising avenue to refine parsing quality. Unlike supervised fine-tuning (SFT), RL encourages the learning of robust, generalizable representations rather than superficial token-level pattern matching~\cite{Logics_parsing_2025}. Existing reward designs generally fall into two categories: VLM-based approaches that employ VLMs as judges to compare rendered predictions against ground truths~\cite{MonkeyOCR_v1_5_2025, TRivia_2025}, and rule-based approaches built on structure-aware metrics like TEDS~\cite{FD_RL_2026, HunyuanOCR_2025, LightOnOCR_2026}. However, as shown in~\cref{fig:RL_related_work}, both exhibit fundamental flaws. VLM-based judges struggle to reliably distinguish fine-grained differences between rendered images, while rule-based metrics may yield misleading rewards when visually identical content admits multiple valid representations (e.g., in formula-intensive tables). Consequently, current reward designs fail to provide stable and reliable optimization signals, leaving the potential of RL for table parsing largely untapped.

Furthermore, progress in table parsing is severely constrained by the limitations of existing benchmarks. Recent document understanding benchmarks~\cite{OCRBench_v2_2026, CC_OCR_2025, Omnidocbench_2025} typically treat table parsing as an auxiliary task, resulting in small evaluation sets with restricted scenario diversity. Consequently, they may fail to adequately capture the complexity of real-world table parsing, where variations in acquisition conditions, complex layouts, and diverse writing styles pose significant challenges to model robustness and generalization~\cite{Wilddoc_2025, DocPTBench_2025}.

To address these critical limitations across task modeling, optimization, and evaluation, we propose a comprehensive framework that rethinks table parsing from a full-stack perspective. At the core of our framework is \textbf{StrucTab}, a 1B-parameter VLM.
At the modeling level (\cref{fig:introduction} (b)), we decompose the parsing process into three complementary subtasks: row--column counting, merged-cell analysis, and final HTML generation. These tasks are initially introduced independently during early pretraining to build atomic skills and are subsequently unified via a sequential reasoning strategy. This step-by-step formulation progressively equips the model with robust table-understanding capabilities.
At the optimization level (\cref{fig:introduction} (c)), we introduce \textbf{Uni-TabRL}, a novel RL framework tailored for table parsing. Building upon GRPO~\cite{Deepseekmath_2024}, it incorporates fine-grained, decomposed rewards, thereby providing more stable and reliable optimization signals than existing VLM- or rule-based approaches, leading to more pronounced performance gains.
Finally, at the evaluation level (\cref{fig:introduction} (d)), we present \textbf{TableVerse-5K}, a large-scale, challenging benchmark spanning diverse real-world scenarios, including photographed, printed, and handwritten tables. Extensive experiments demonstrate that StrucTab achieves state-of-the-art performance across all evaluated public benchmarks and significantly outperforms all the baselines on TableVerse-5K.

We summarize our main contributions as follows:
\begin{itemize}[
        label=\raisebox{0.5ex}{\tiny$\bullet$},
        leftmargin=1em,
        itemsep=2pt, % 项与项之间的间距
        parsep=0pt, % 段落之间的间距
        topsep=0pt, % 列表与上下文的间距
        partopsep=0pt % 在新段落开始时的额外间距
    ]
    \item \textbf{Modeling.} We present \textbf{StrucTab}, a structure-aware VLM that achieves state-of-the-art performance across multiple benchmarks. Through explicitly incorporating human-inspired structural reasoning, it substantially improves robustness in complex tables over standard black-box approaches.
    \item \textbf{Optimization.} We introduce \textbf{Uni-TabRL}, a unified RL framework for table parsing that leverages fine-grained, decomposed rewards to ensure stable and reliable optimization signals, yielding further improvements.
    \item \textbf{Evaluation.} We curate \textbf{TableVerse-5K}, a large-scale, challenging table parsing benchmark encompassing diverse real-world table scenarios to facilitate comprehensive evaluation of model generalization.
\end{itemize}

%% file: figure_code/introduction.tex
\begin{figure}[t!]
    \centering
    \includegraphics[width=\linewidth]{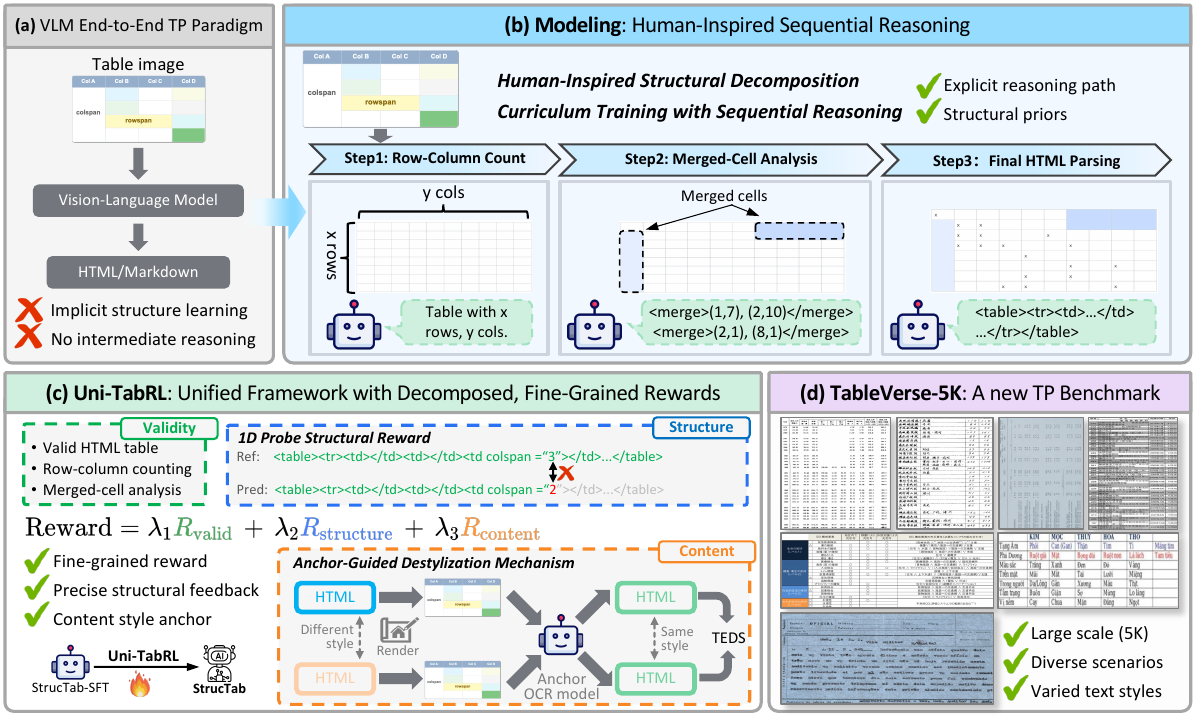}
    \caption{
        \textbf{StrucTab:} A table parsing framework that overcomes the limitations of
        (a) current VLM-based end-to-end paradigms through three key innovations:
        \textbf{(b) Modeling} via human-inspired sequential reasoning;
        \textbf{(c) Optimization} with \textbf{Uni-TabRL} using decomposed, fine-grained rewards;
        and \textbf{(d) Evaluation} on \textbf{TableVerse-5K}, a large-scale, challenging benchmark of diverse real-world tables.
    }
    \label{fig:introduction}
\end{figure}

%% file: figure_code/RL_related_work.tex
\begin{figure}[t!]
    \centering
    \includegraphics[width=\linewidth]{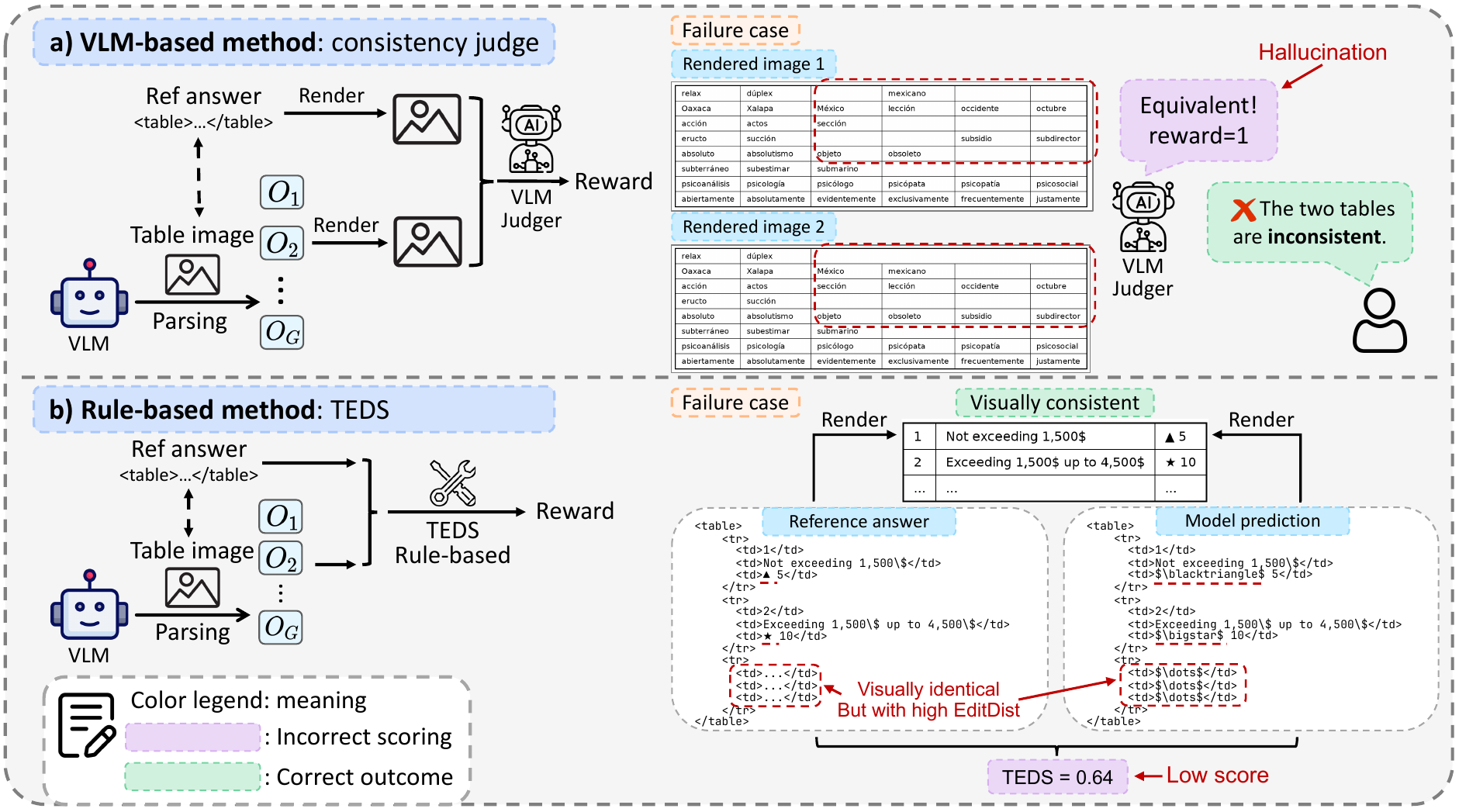}
    \caption{
        \textbf{Limitations of existing RL reward designs for table parsing.}
        Top: VLM-based methods rely on VLM judges to compare rendered predictions with references, but may introduce noisy supervision due to hallucinations or misalignment.
        Bottom: Rule-based methods measure markup similarity and can be misleading when visually identical tables admit multiple valid expressions, especially in formula-intensive cases.
        These issues lead to unstable and ambiguous optimization signals.
    }
    \label{fig:RL_related_work}
\end{figure}

%% file: section/2_background.tex
\section{Background}
\label{sec:background}

\input{section/2_background/2_1_preliminaries}
\input{section/2_background/2_2_related_work}

%% file: section/2_background/2_1_preliminaries.tex
\subsection{Preliminaries}
\label{subsec:preliminaries}

\mypara{TEDS.}
Tree-Edit-Distance-based Similarity (TEDS)~\cite{Image_table_2020} is widely used in table parsing tasks to measure differences between prediction and ground truth, jointly evaluating structural alignment and cell content correctness~\cite{OCRBench_v2_2026, CC_OCR_2025, Omnidocbench_2025}. As in~\cref{eq:teds}, $T_{\text{pred}}$ and $T_{\text{gt}}$ denote the predicted and ground-truth HTML trees, with $|T|$ representing the node count of $T$. The edit distance $\text{EditDist}(T_{\text{pred}}, T_{\text{gt}})$ is the minimum number of node operations required to transform $T_{\text{pred}}$ into $T_{\text{gt}}$:

\input{equation/TEDS}

Instead, TEDS-S (structure-only TEDS) applies the same formulation but ignores textual content, evaluating the structural accuracy of prediction alone.

\mypara{GRPO.}
Group Relative Policy Optimization (GRPO)~\cite{Deepseekmath_2024} has been widely adopted in RL as a critic-free algorithm that directly optimizes policies using group-normalized rewards from sampled responses. For each input $x$, multiple outputs $\{o_1,o_2,\ldots,o_G\}$ are sampled from the current policy $\theta_{\mathrm{old}}$, and rewards $R_i$ are computed. The group-normalized advantage for the $i$-th response is:

\input{equation/GRPO_advantage}

The policy model $\pi_{\theta}$ is then optimized by maximizing:

\input{equation/GRPO_loss}

\noindent
where $\epsilon$ is a hyperparameter for clipping the policy ratio.

%% file: equation/TEDS.tex
\vspace{-0.3em}
\begin{equation}
    \text{TEDS}(T_{\text{pred}}, T_{\text{gt}})
    =
    1 - \frac{\text{EditDist}(T_{\text{pred}}, T_{\text{gt}})}{\max(|T_{\text{pred}}|, |T_{\text{gt}}|)} .
    \label{eq:teds}
\end{equation}
\par\nobreak\vspace{-0.1em}

%% file: equation/GRPO_advantage.tex
\vspace{-0.4em}
\begin{equation}
    A_{i}
    =
    \frac{R_{i} - \mathrm{mean}(\{R_j\}_{j=1}^G)}{\mathrm{std}(\{R_{j}\}_{j=1}^G)} .
    \label{eq:GRPO_advantage}
\end{equation}
\par\nobreak\vspace{-0.1em}

%% file: equation/GRPO_loss.tex
\vspace{-0.4em}
\begin{equation}
    \begin{aligned}
        J_{\text{GRPO}}(\theta)
        =
         &
        \mathbb{E}_{(x,y)\sim D,\,\{o_i\}_{i=1}^{G}\sim\,\pi_{\theta_{\text{old}}}(\cdot|x)}
        \\[-0.2em]
         &
        \left[
            \frac{1}{G}\!\sum_{i=1}^{G} \min\!\left(\!\frac{\pi_\theta(o_i|x)}{\pi_{\theta_{\text{old}}}(o_i|x)}A_{i},\text{clip}\!\left(\frac{\pi_\theta(o_i|x)}{\pi_{\theta_{\text{old}}}(o_i|x)},1-\epsilon,1+\epsilon\!\right)A_{i}\!\right)\!
        \right],
    \end{aligned}
    \label{eq:GRPO_loss}
\end{equation}
\par\nobreak\vspace{-0.1em}

%% file: section/2_background/2_2_related_work.tex
\subsection{Related Work}
\label{subsec:related_work}

\mypara{Table Parsing.}
Traditional table parsing methods primarily adopted multi-stage pipelines, decomposing the task into structure recognition and OCR~\cite{PP_OCRv2_2021, Nougat_2024, Vary_2024, Mineru_2024, PaddleOCR_3_2025, Marker_2025, Tsrformer_2022, Split_2022}. While effective on clean documents, these approaches suffer from error propagation and limited robustness in real-world scenarios~\cite{parsing_table_wild_2021}. To mitigate these issues, image-to-markup approaches reformulate parsing as an end-to-end sequence generation task, directly converting table images into structured formats such as HTML~\cite{Tableformer_2022, VAST_2023, UniTable_2024}. Modern VLMs have further advanced this paradigm through unified modeling and large-scale pretraining~\cite{GOT_OCR_2024, Dots_ocr_2025, Logics_parsing_2025, olmOCR_2025}. However, most current end-to-end methods optimize for the final HTML sequence directly, overlooking explicit intermediate structural reasoning. This holistic optimization may fail to capture intricate table structures robustly.

\mypara{RL for Table Parsing.}
Recent studies explore RL for table parsing via task-specific rewards. FD-RL~\cite{FD_RL_2026} adopts rule-based TEDS, which may be misleading when semantically equivalent content (e.g., formulas) admits multiple syntactic forms. MonkeyOCR-v1.5~\cite{MonkeyOCR_v1_5_2025} uses a VLM-based judge to assess visual consistency between rendered predictions and ground truth, but limited VLM sensitivity to fine-grained details introduces noisy feedback. TRivia~\cite{TRivia_2025} employs auxiliary QA tasks as reinforcement signals, focusing on local regions rather than global structural correctness. Collectively, these works highlight the potential of RL for table parsing, while revealing the challenge of designing stable, fine-grained rewards for holistic table understanding~\cite{Table_R1_2025, TableGPT_R1_2025, Reasoning_Table_2025, Uni_DPO_2025}.

\mypara{Table Parsing Benchmark.}
Table parsing is typically evaluated on small subsets of document understanding benchmarks, such as OmniDocBench~\cite{Omnidocbench_2025}, CC-OCR~\cite{CC_OCR_2025}, and OCRBench v2~\cite{OCRBench_v2_2026}. As parsing serves only as an auxiliary subtask in these suites, the samples are limited in scale and structural diversity. Consequently, they may fail to fully reflect real-world complexities such as intricate merged cells, diverse acquisition conditions, and handwritten content, highlighting the need for more dedicated and comprehensive benchmarks.

%% file: section/3_StrucTab.tex
\section{StrucTab: Framework and Training Strategy}
\label{sec:StrucTab}

\input{figure_code/optimization_framework}

In this section, we present the structured reasoning and optimization framework of StrucTab. We first introduce the human-inspired structural decomposition for table parsing in~\cref{subsec:structural_decomposition}, followed by a curriculum training strategy with sequential reasoning in~\cref{subsec:curriculum_training}. Finally, we describe the Uni-TabRL optimization framework in~\cref{subsec:Uni_TabRL}. The overall optimization pipeline is illustrated in~\cref{fig:optimization_framework}.

\input{section/3_StrucTab/3_1_structural_decomposition}
\input{section/3_StrucTab/3_2_curriculum_training}
\input{section/3_StrucTab/3_3_Uni_TabRL}

%% file: figure_code/optimization_framework.tex
\begin{figure}[t!]
    \centering
    \includegraphics[width=\linewidth]{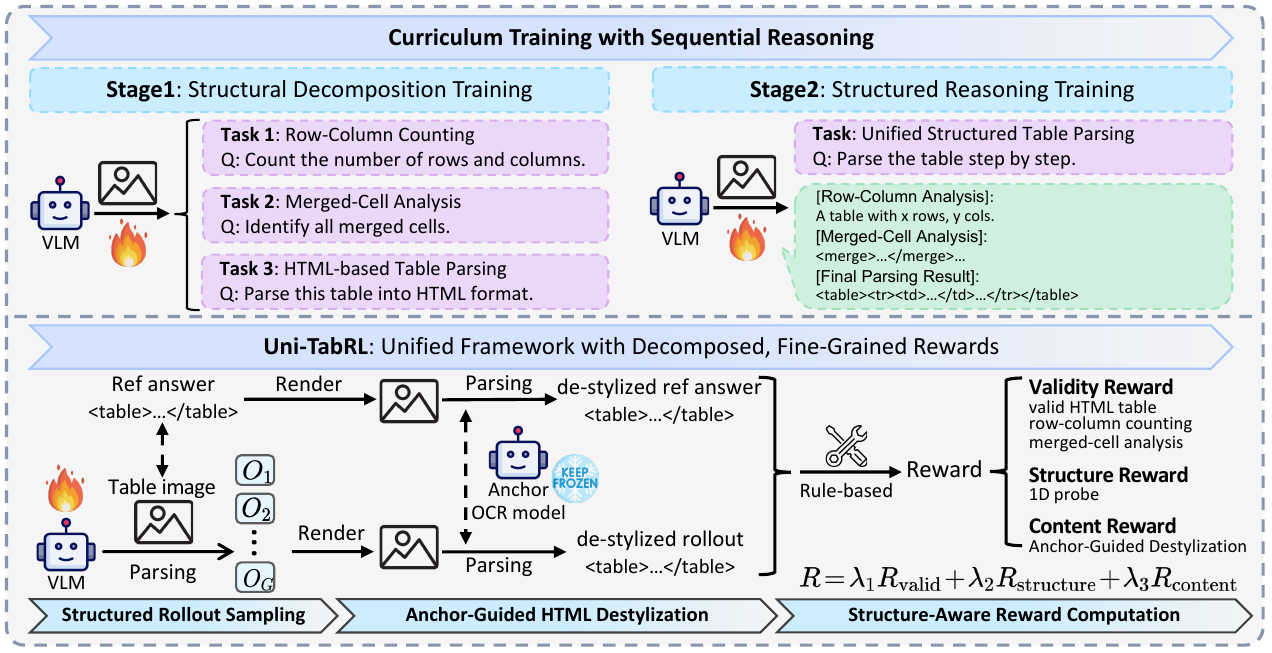}
    \vspace{-1.2em}
    \caption{
        \textbf{Structured Optimization Framework for StrucTab.}
        Top: Curriculum training progressively integrates atomic structural skills into unified reasoning.
        Bottom: Uni-TabRL enables stable optimization via fine-grained, decomposed rewards.
    }
    \label{fig:optimization_framework}
\end{figure}

%% file: section/3_StrucTab/3_1_structural_decomposition.tex
\subsection{Human-Inspired Structural Decomposition}
\label{subsec:structural_decomposition}

Inspired by the explicit reasoning process of humans, we hypothesize that intermediate structural signals play an important role in table understanding but are not explicitly modeled in current end-to-end training. To verify this, we conduct a training-free preliminary study by injecting structural cues during parsing. As shown in~\cref{subfig:hint_result}, providing row--column counts or merged-cell information as hints consistently improves table parsing accuracy. This result suggests that intermediate structural signals are indeed underutilized in existing models.

Motivated by this observation, we reformulate table parsing as structured reasoning and introduce intermediate objectives into training. Specifically, we decompose the task into two complementary subtasks: (a) row--column counting and (b) merged-cell analysis. These components inject explicit structural priors before final HTML decoding, guiding the model toward more robust and stable table understanding. Both subtasks are lightweight and directly derived from standard HTML annotations, enabling scalable supervision without additional manual labeling. Details are provided in the Appendix.

\mypara{Row--Column Counting.}
This subtask targets global structural perception by estimating the number of rows and columns in a table. Given a table image, the model is prompted:
\emph{``Count the number of rows and columns in this table.''}
The target is formatted as a concise description (e.g., \emph{``$x$ rows and $y$ columns''}). By predicting global dimensions before detailed parsing, the model develops a holistic understanding of table layout.

\mypara{Merged--Cell Analysis.}
This subtask models local structural dependencies by identifying merged cells. Using the \texttt{rowspan} and \texttt{colspan} attributes in HTML annotations, we derive merged regions and represent them in a structured tokenized format: \texttt{<merge>(r\textsubscript{1},c\textsubscript{1}),(r\textsubscript{2},c\textsubscript{2})</merge>}, indicating a cell spanning from $(r_1, c_1)$ to $(r_2, c_2)$. Merged regions are extracted row by row and concatenated in a fixed order to form the target sequence. The prompt is:
\emph{``Identify all merged cells in this table.''}
Explicit supervision of merged relationships equips the model with localized structural awareness.

\input{figure_code/training_free_demo}

%% file: figure_code/training_free_demo.tex
\begin{figure}[t!]
    \centering
    \begin{subfigure}[t]{0.59\linewidth}
        \centering
        \includegraphics[width=\linewidth]{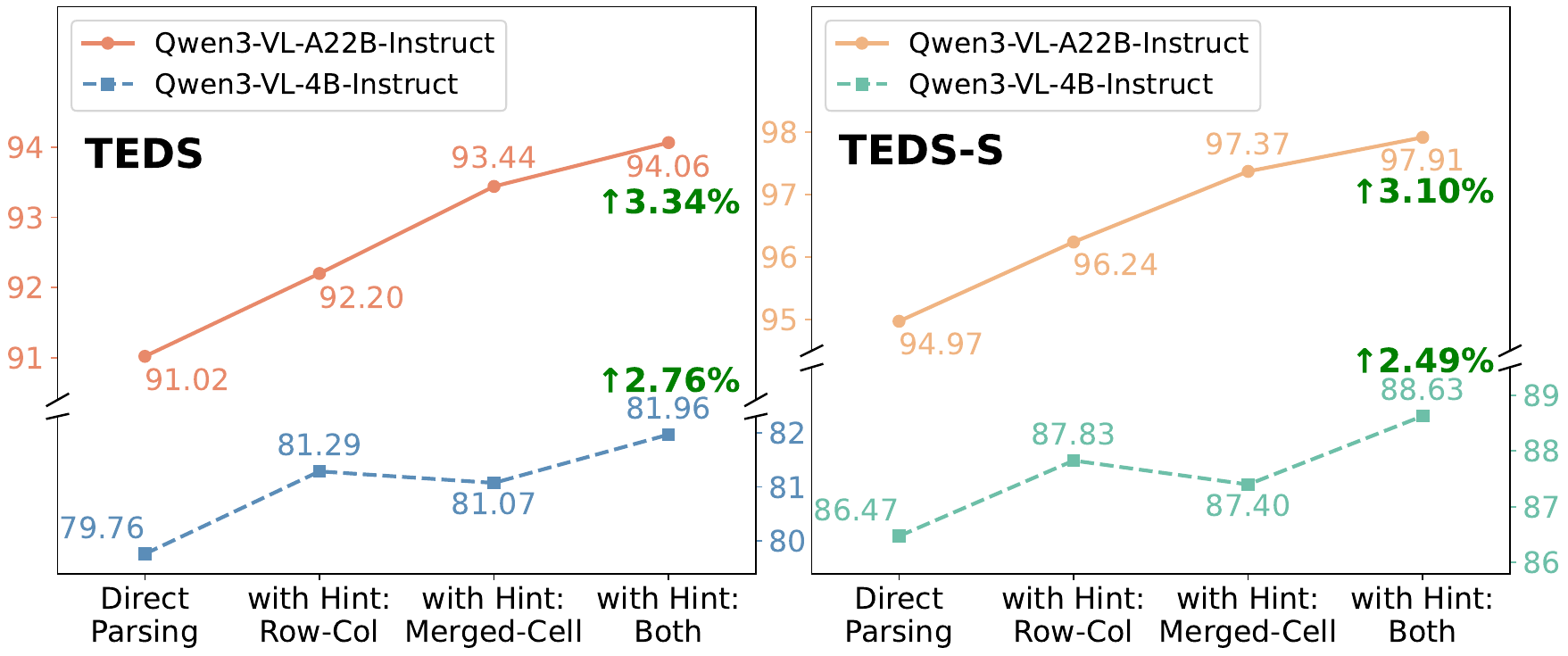}
        \vspace{-1.0em}
        \caption{Impact of explicit structural cues}
        \label{subfig:hint_result}
    \end{subfigure}
    \begin{subfigure}[t]{0.39\linewidth}
        \centering
        \includegraphics[width=\linewidth]{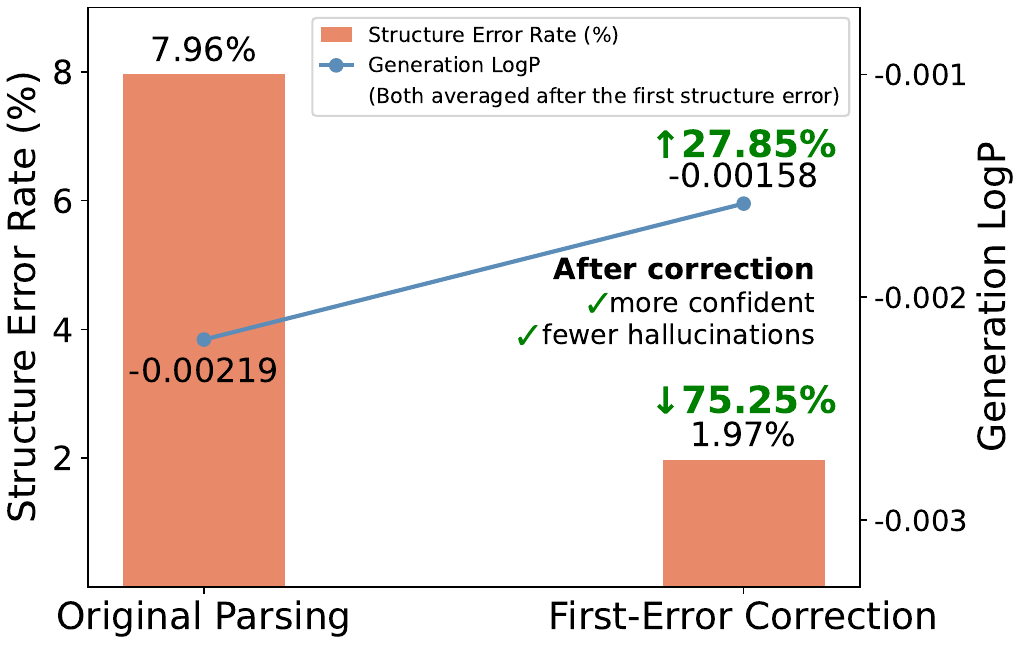}
        \vspace{-1.0em}
        \caption{Impact of early error correction}
        \label{subfig:error_rate_and_logp_comparison}
    \end{subfigure}
    \caption{
        \textbf{(a)} Providing structural cues (e.g., row--column counts) explicitly improves parsing performance on OmniDocBench.
        \textbf{(b)} Correcting the first structural error reduces subsequent structural errors while improving downstream generation confidence.
    }
    \label{fig:structural_motivation}
\end{figure}

%% file: section/3_StrucTab/3_2_curriculum_training.tex
\subsection{Curriculum Training with Sequential Reasoning}
\label{subsec:curriculum_training}

To bridge structural decomposition and end-to-end parsing, we design a curriculum training strategy with sequential reasoning. In the first stage of pretraining, row--column counting, merged-cell analysis, and HTML parsing are optimized as three independent question--answer tasks. This stage focuses on learning atomic structural perception and parsing skills in isolation.

In the second stage, we integrate these subtasks into a unified reasoning sequence that follows the human-inspired perception order. Specifically, we construct single-round dialogues requiring the model to sequentially generate (a) row--column counts, (b) merged-cell analysis results, and (c) the final HTML representation within one coherent response. This sequential formulation imposes an explicit reasoning path, encouraging the model to condition its final parsing decisions on intermediate structural predictions.

Through curriculum-based integration and sequential supervision, StrucTab learns step-by-step structured reasoning, leading to more robust and interpretable table parsing behavior compared to conventional direct end-to-end training.

%% file: section/3_StrucTab/3_3_Uni_TabRL.tex
\subsection{Uni-TabRL}
\label{subsec:Uni_TabRL}

\input{figure_code/RL_method_details}

RL is effective in improving document understanding models by designing task-specific rewards~\cite{Infinity_parser_2025, Logics_parsing_2025}, but existing reward signals are often noisy or ambiguous for structural reasoning ability learning, thus remaining suboptimal for robust table parsing. To supply fine-grained and reliable supervision for RL, we decompose the overall reward into three complementary components that separately supervise \textit{generation validity}, \textit{structural correctness}, and \textit{content fidelity}:

\input{equation/RL_reward}

\noindent
where $\lambda_{1}$, $\lambda_{2}$, and $\lambda_{3}$ are hyperparameters balancing the contributions of each component. Details are provided in the Appendix.

\mypara{Validity Reward.}
To guarantee the feasibility of generated outputs, we introduce this component as a hard gating mechanism. Given the strict syntax requirements of HTML, any failure to produce a complete and valid table or correct prerequisite outputs (incorrect row--column counting or merged-cell analysis) results in a zero validity reward; otherwise, $R_{\text{valid}}=1$. This binary constraint forces the model to master basic structural validity and task decomposition as a foundation for subsequent fine-grained optimization.

\mypara{Structure Reward.}
To provide precise feedback on the structural correctness, we move beyond holistic metrics like TEDS-S as a reward, which yield a single aggregated similarity score without identifying where structural errors occur or how they propagate during decoding. Such non-decomposable signals fail to provide fine-grained, step-level guidance for autoregressive generation. Our preliminary study in~\cref{subfig:error_rate_and_logp_comparison} reveals that early correction of structural deviations is crucial for stabilizing long-form decoding and leads to more reliable table generation. Thus, we propose a \textbf{1D Probe structural reward} tailored for autoregressive models. As shown in~\cref{subfig:1D_probe_structural_reward}, by linearizing both the predicted and reference tables into cell sequences, we compute the reward as the ratio of correctly matched cells before the first mismatch. This mechanism explicitly penalizes early structural divergence and encourages the model to maintain a correct structural trajectory, aligning the optimization signal with the sequential generation process.

\mypara{Content Reward.}
While standard TEDS is widely used to evaluate content accuracy, it often penalizes valid outputs that differ stylistically from the reference (e.g., visually equivalent formulas), introducing noise into reward signals. Modern OCR models are trained extensively on synthetic and rendered document images, enabling high-precision and deterministic parsing on standardized inputs. Leveraging this property, we propose \textbf{Anchor-Guided Destylization} to resolve this ambiguity. Specifically, both the reference and the model's output are rendered into standardized images (illustrated in~\cref{subfig:render_pipeline}) and then re-parsed by a frozen OCR model acting as a style anchor. This process normalizes stylistic variations into a unified canonical format. By computing TEDS between these de-stylized representations, we obtain a robust, style-invariant reward that focuses purely on semantic content fidelity.

By combining validity constraints, localized structural feedback, and style-robust content supervision, our decomposed reward design yields stable and informative signals, effectively guiding the model toward robust table parsing.

%% file: figure_code/RL_method_details.tex
\begin{figure}[t!]
    \centering
    \begin{subfigure}[b]{\linewidth}
        \includegraphics[width=\linewidth]{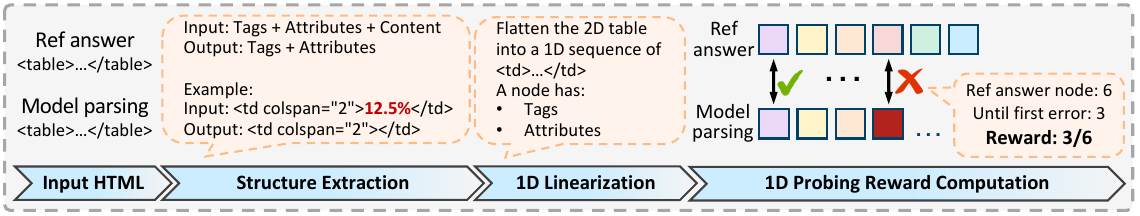}
        \vspace{-1.2em}
        \caption{1D Probe structural reward}
        \label{subfig:1D_probe_structural_reward}
    \end{subfigure}
    \begin{subfigure}[b]{\linewidth}
        \includegraphics[width=\linewidth]{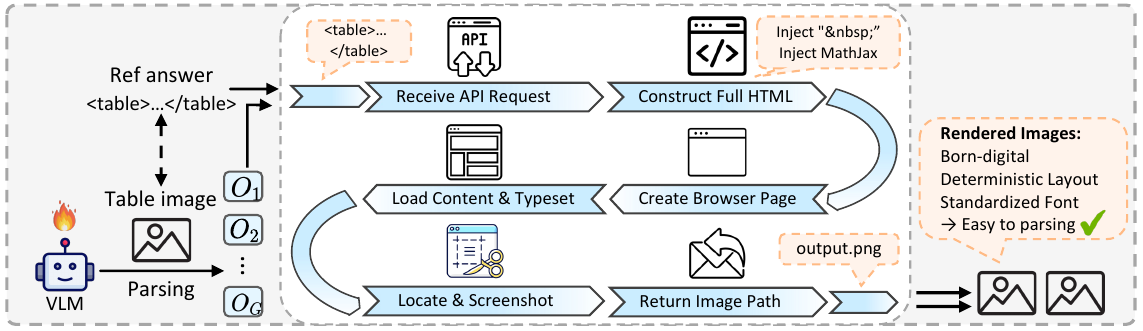}
        \vspace{-1.2em}
        \caption{Table image render pipeline}
        \label{subfig:render_pipeline}
    \end{subfigure}
    \vspace{-1.5em}
    \caption{
        \textbf{Uni-TabRL Implementation Pipeline.}
        Details are in the Appendix.
    }
    \vspace{-0.5em}
    \label{fig:RL_method_details}
\end{figure}

%% file: equation/RL_reward.tex
\vspace{-0.4em}
\begin{equation}
    R
    = \lambda_{1} R_{\text{valid}}
    + \lambda_{2} R_{\text{structure}}
    + \lambda_{3} R_{\text{content}}\, ,
    \label{eq:RL_reward}
\end{equation}
\par\nobreak\vspace{-0.2em}

%% file: section/4_TableVerse.tex
\section{TableVerse-5K Benchmark}
\label{subsec:TableVerse}

For more comprehensive real-world evaluation, we construct \textbf{TableVerse-5K}, a large-scale benchmark featuring diverse table sources and high-quality annotations. Comparison with existing benchmarks is presented in~\cref{tab:benchmark_comparison}, and the construction pipeline is shown in~\cref{fig:benchmark_pipeline}. Details are provided in the Appendix.

\mypara{Data Collection.}
All table images in TableVerse-5K are manually collected from diverse real-world sources, including academic papers, posters, reports, and technical documents~\cite{peng2026chartarena}. The tables are manually cropped to ensure accurate localization. The dataset covers both Chinese and English, spanning printed tables, photographed tables, and handwritten content, thereby reflecting the diversity and complexity of practical scenarios.

\mypara{Annotation Pipeline.}
To ensure both efficiency and annotation quality, we adopt a hybrid annotation strategy that combines multi-VLM-based annotation with rigorous human verification. Specifically, we first obtain candidate parsing results using three strong models: PaddleOCR-VL~\cite{PaddleOCR_VL_2025}, MinerU2.5~\cite{Mineru_2_5_2025}, and HunyuanOCR~\cite{HunyuanOCR_2025}. Human annotators then select the most complete prediction as the initial draft and refine it by carefully correcting structural elements (e.g., layout and merged cells) as well as cell contents (including text and formulas). The refined annotations are subsequently reviewed by a second annotator for further verification and correction, forming a multi-stage quality control process~\cite{peng2026chartarena, li2026chronicles}.

\mypara{Dataset Summary.}
Following this pipeline, we construct TableVerse-5K, a benchmark comprising 5K high-quality annotated tables. Compared with existing benchmarks, it offers greater scenario diversity, richer structural variations, and stricter annotation standards, making it better suited for evaluating the robustness and generalization of table parsing models in real-world settings. A detailed comparison and additional visualizations are provided in the Appendix.

\input{table/benchmark_comparison}
\input{figure_code/benchmark_pipeline}

%% file: table/benchmark_comparison.tex
\begin{table}[t!]
    \centering
    \captionsetup{font={small}}
    \caption{
        \textbf{Comparison of Table Parsing Benchmarks.} Our TableVerse-5K is the largest, most diverse, and structurally complex benchmark for real-world table parsing.
    }
    \vspace{-0.6em}
    {\renewcommand{\arraystretch}{0.95}
        \setlength{\tabcolsep}{5pt}
        \resizebox{\columnwidth}{!} {%
            \begin{tabular}{l c c c cc ccc}
                \toprule
                \multirow{2}{*}[-0.4mm]
                {\textbf{Benchmark}}                                    &
                \multirow{2}{*}[-0.4mm]
                {\textbf{Release Date}}                                 &
                \multirow{2}{*}[-0.4mm]
                {\textbf{Size}}                                         &
                \multirow{2}{*}[-0.4mm]
                {\textbf{\makecell{Merged Cells \nextline Avg. Count}}} &
                \multicolumn{2}{c}{\textbf{Avg. Dim.}}                  &
                \multicolumn{3}{c}{\textbf{Scenario}}
                \\
                \cmidrule(lr){5-6}
                \cmidrule(lr){7-9}
                                                                        &         &                &               & Rows           & Cols          & Scan   & Photo (Printed) & Photo (Handwritten) \\
                \midrule
                OmniDocBench~\cite{Omnidocbench_2025}                   & 2024.12 & 512            & 1.34          & 10.04          & 5.40          & \cmark & \xmark          & \xmark              \\
                CC-OCR~\cite{CC_OCR_2025}                               & 2024.12 & 300            & 4.63          & \textbf{18.99} & \textbf{7.37} & \cmark & \cmark          & \cmark              \\
                OCRBench v2~\cite{OCRBench_v2_2026}                     & 2025.01 & 700            & 1.37          & 8.19           & 4.55          & \cmark & \cmark          & \xmark              \\
                \midrule
                \rowcolor{table_ours}
                \textbf{TableVerse-5K (Ours)}                           & 2026.06 & \textbf{5,000} & \textbf{5.10} & 14.39          & 7.13          & \cmark & \cmark          & \cmark              \\
                \bottomrule
            \end{tabular}
        }
    }
    \label{tab:benchmark_comparison}
\end{table}

%% file: figure_code/benchmark_pipeline.tex
\begin{figure}[t!]
    \centering
    \includegraphics[width=\linewidth]{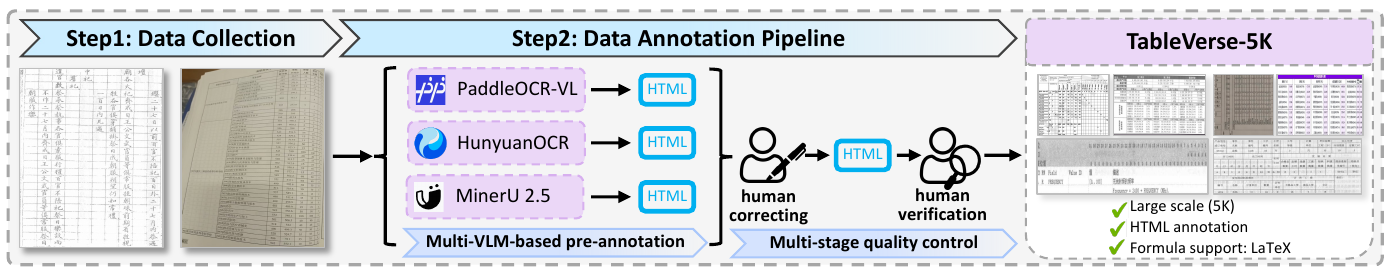}
    \vspace{-1.6em}
    \caption{
        \textbf{TableVerse-5K Construction Pipeline.}
    }
    \vspace{-0.5em}
    \label{fig:benchmark_pipeline}
\end{figure}

%% file: section/5_experiments.tex
\section{Experiments}
\label{sec:experiments}

In this section, we conduct comprehensive experiments to demonstrate the superior performance of StrucTab. We detail the experimental settings in~\cref{subsec:experimental_settings}, report the results in~\cref{subsec:experimental_results}, and provide ablation studies in~\cref{subsec:ablation_study}. Finally, we present qualitative visualizations in~\cref{subsec:qualitative}.

\subsection{Experimental Settings}
\label{subsec:experimental_settings}

\mypara{Training Pipeline.}
We adopt HunyuanOCR~\cite{HunyuanOCR_2025} as the vision-language backbone, as it currently represents one of the strongest lightweight OCR models. Full-parameter fine-tuning is performed throughout all training stages.
The training pipeline consists of three stages: pretraining, supervised fine-tuning (SFT), and RL. All experiments are conducted on 16 NVIDIA H20 GPUs. Additional implementation details are provided in the Appendix.

\textbf{(a) Pretraining.}
We conduct pretraining in two progressive stages on a mixture of 6M synthetic table samples and publicly available datasets~\cite{TabRecSet_2023, SEMv2_2024, FinTabNet_2021, Image_table_2020, Tableformer_2022}. In the first stage, row--column counting, merged-cell analysis, and HTML parsing are treated as three independent objectives. The model is trained for one epoch with a learning rate of $4 \times 10^{-5}$. In the second stage, the three tasks are unified into single-round dialogues with a fixed reasoning order. This stage is trained for one additional epoch.

\textbf{(b) SFT.}
For downstream adaptation, we construct a 130K-sample high-quality dataset comprising 50K samples from public datasets and 80K manually annotated HTML tables. Following the second pretraining stage, we use 100K samples for SFT, training for two epochs at $2 \times 10^{-5}$.

\textbf{(c) RL.}
Finally, we perform RL on the remaining 30K samples using the proposed Uni-TabRL framework to further optimize parsing performance.

\mypara{Evaluation Benchmarks.}
We evaluate on the table parsing subsets of three widely used benchmarks: OmniDocBench~\cite{Omnidocbench_2025}, CC-OCR~\cite{CC_OCR_2025}, and OCRBench v2~\cite{OCRBench_v2_2026}, covering digital, scanned, and photographed tables of varying complexity.
We further assess performance on our proposed TableVerse-5K, a larger and more challenging benchmark with substantial real-world diversity.

\mypara{Baselines.}
We compare our method against representative baselines across three categories:
(a) \textbf{Expert table parsing (TP) models}, dedicated models designed explicitly for table structure recognition and parsing;
(b) \textbf{General-purpose VLMs}, capable of table parsing via unified multimodal modeling;
and (c) \textbf{Document parsing VLMs}, optimized for holistic document parsing.
For each category, we select strong and representative models for comparison to ensure a comprehensive evaluation.

\subsection{Experimental Results}
\label{subsec:experimental_results}

\input{table/experiment_results}

The quantitative results in~\cref{tab:experiment_results} show that StrucTab achieves state-of-the-art performance across all three public benchmarks and TableVerse-5K, consistently outperforming the baselines by a clear margin. Notably, despite its efficient 1B-parameter backbone, StrucTab significantly surpasses massive closed-source VLMs, exceeding Gemini 2.5 Pro and GPT-5 by 7.13\% and 19.14\% in average TEDS, respectively. This highlights that explicit structural decomposition handles complex layouts much more effectively than the proprietary models.

Furthermore, StrucTab demonstrates clear superiority over existing RL-based table parsers, which often suffer from noisy or ambiguous reward signals. Our approach outperforms FD-RL and TRivia-3B by 13.24\% and 7.92\% in average TEDS, respectively. Crucially, the Uni-TabRL framework alone drives a 2.92\% TEDS increase over our supervised baseline (StrucTab-SFT). Overall, these results validate our design: combining intermediate structural reasoning with stable, fine-grained RL optimization significantly enhances parsing robustness.

\subsection{Ablation Study}
\label{subsec:ablation_study}

\mypara{Effectiveness of Human-Inspired Structural Decomposition.}
We conduct ablation studies during both pretraining and SFT to evaluate the impact of our human-inspired structural decomposition, and summarize the results in~\cref{tab:SFT_ablation}. Starting from an end-to-end table-to-HTML parsing baseline (a), we introduce row--column counting and merged-cell analysis as auxiliary supervision signals in settings (b)--(d). Notably, in these settings, tasks are trained independently without sequential reasoning. Results show that incorporating either structural subtask consistently improves performance, and their combination yields further gains. This demonstrates that explicit supervision of intermediate structural properties provides valuable inductive biases, enabling the model to learn robust representations beyond merely fitting the final HTML sequence.

\input{table/SFT_ablation}

\mypara{Effectiveness of Curriculum Training and Sequential Reasoning.}
We further investigate the efficacy of our curriculum training strategy, with results reported in~\cref{tab:SFT_ablation}. Comparing setting (d) (independent tasks) with setting (f) (single-round unified sequential reasoning), we observe that integrating subtasks into a coherent reasoning chain significantly boosts performance. This indicates that enforcing an explicit reasoning path allows the model to better condition final table reconstruction on intermediate structural predictions, rather than treating them as isolated auxiliary objectives. Moreover, to validate the necessity of the two-stage curriculum, we conduct a controlled experiment (e) where the first-stage pretraining of independent subtasks is removed, training directly with sequential reasoning. As shown in (e) vs. (f), bypassing atomic skill learning leads to inferior performance. This highlights that mastering independent structural capabilities before integrating them into a unified reasoning process is crucial for stable optimization and robust table understanding.

\mypara{Effectiveness of Uni-TabRL.}
As shown in~\cref{tab:Uni_TabRL}, variant (a) outperforms the SFT baseline, demonstrating the effectiveness of GRPO-based reinforcement learning for table parsing. Moreover, (c) achieves better performance than both (a) and (b), verifying the advantage of the Anchor-Guided Destylization mechanism, while the inferior results of (b) suggest that VLM-based consistency judges may introduce noisy or unreliable supervision signals that hinder optimization. In addition, (d) improves over (a), confirming the effectiveness of the proposed 1D Probe mechanism as a structure reward. Overall, combining Anchor-Guided Destylization with the 1D Probe structural reward yields the best performance, leading to the final RL model consistently outperforming all other configurations.

\input{table/Uni_TabRL}

\mypara{1D Probe Structural Reward Mitigates Degenerate Repetition.}
The repetition issue arises when the model fails to generate the closing `</table>' token, leading to an infinite repetition loop during autoregressive decoding. This behavior severely compromises the practicality and deployability of table parsing systems and has therefore drawn significant attention in recent studies~\cite{olmOCR_2_2025, Mineru_2_5_2025, DS_OCR_2_2026, LightOnOCR_2026}. As shown in~\cref{tab:RL_on_repetition}, both SFT and RL substantially reduce the repetition rate, while the 1D Probe structural reward further mitigates this issue, demonstrating its effectiveness in guiding autoregressive learning.

\input{table/RL_on_repetition}

\subsection{Qualitative Visualization on TableVerse-5K}
\label{subsec:qualitative}

\input{figure_code/qualitative_result}

\cref{fig:qualitative_result} shows representative examples from TableVerse-5K comparing the baseline (c) HunyuanOCR with (d) StrucTab-SFT and (e) StrucTab, where the baseline often produces structural inconsistencies or content errors when directly generating HTML tables. After introducing structural decomposition, StrucTab-SFT improves parsing quality by explicitly reasoning over row--column counts and merged cells before table reconstruction.
With further optimization using Uni-TabRL, StrucTab achieves more accurate structural alignment and content consistency, producing predictions that more closely match the ground truth. These visual results are consistent with the experimental improvements observed in our quantitative evaluations, demonstrating the effectiveness of both the structured reasoning paradigm and the proposed Uni-TabRL.

%% file: table/experiment_results.tex
\begin{table}[t!]
    \centering
    \captionsetup{font={small}}
    \caption{\textbf{Main experimental results.}
        StrucTab achieves the best performance.
    }
    \vspace{-0.4em}
    {\renewcommand{\arraystretch}{0.95}
        \setlength{\tabcolsep}{5pt}
        \resizebox{\columnwidth}{!} {%
            \begin{tabular}{cll cccccccccc}
                \toprule
                \multirow{2}{*}[-0.4mm]
                {\textbf{\makecell{Model \nextline Type}}}   &
                \multirow{2}{*}[-0.4mm]
                {\textbf{Model}}                             &
                \multirow{2}{*}[-0.4mm]
                {\textbf{\makecell{Release \nextline Date}}} &
                \multicolumn{2}{c}{\textbf{OmniDocBench}}    &
                \multicolumn{2}{c}{\textbf{CC-OCR}}          &
                \multicolumn{2}{c}{\textbf{OCRBench v2}}     &
                \multicolumn{2}{c}{\textbf{TableVerse-5K}}   &
                \multicolumn{2}{c}{\textbf{Average}}
                \\
                \cmidrule(lr){4-5}
                \cmidrule(lr){6-7}
                \cmidrule(lr){8-9}
                \cmidrule(lr){10-11}
                \cmidrule(lr){12-13}
                                                             &                                            &         & TEDS           & TEDS-S         & TEDS           & TEDS-S         & TEDS           & TEDS-S         & TEDS           & TEDS-S         & TEDS           & TEDS-S         \\
                \midrule
                \multirow{2}{*}{\makecell{Expert TP \nextline models}}
                                                             & UniTable~\cite{UniTable_2024}              & 2024.03 & 82.76          & 89.82          & 57.84          & 70.47          & 67.73          & 78.65          & 48.55          & 78.65          & 53.73          & 79.15          \\
                                                             & TRivia-3B~\cite{TRivia_2025}               & 2025.12 & 91.60          & 95.01          & 84.90          & 90.17          & 90.76          & 94.03          & 78.15          & 85.41          & 80.87          & 87.31          \\
                \midrule
                \multirow{8}{*}{\makecell{General \nextline Purpose \nextline VLMs}}
                                                             & GPT-4o~\cite{GPT_4o_2024}                  & 2024.05 & 78.27          & 84.56          & 66.98          & 79.04          & 70.51          & 79.55          & 63.62          & 76.41          & 65.67          & 77.51          \\
                                                             & GPT-5~\cite{GPT_5_2025}                    & 2025.08 & 84.91          & 89.91          & 63.25          & 74.09          & 79.91          & 88.69          & 67.04          & 78.96          & 69.65          & 80.64          \\
                                                             & Qwen2.5-VL-72B-Ins.~\cite{Qwen2.5-VL}      & 2025.02 & 87.85          & 91.80          & 81.22          & 86.48          & 81.33          & 86.58          & 75.23          & 82.65          & 77.15          & 83.97          \\
                                                             & InternVL3.5-A28B~\cite{InternVL3_5_2025}   & 2025.08 & 86.03          & 90.53          & 62.87          & 69.52          & 79.50          & 85.81          & 76.08          & 84.96          & 76.62          & 84.78          \\
                                                             & Qwen3-VL-A22B-Ins.~\cite{Qwen3-VL}         & 2025.10 & 91.02          & 94.97          & 80.98          & 86.19          & 84.12          & 88.15          & 78.26          & 84.23          & 80.02          & 85.59          \\
                                                             & Seed-1.8 (no-think)~\cite{Seed_1_8_2025}   & 2025.12 & 89.09          & 94.37          & 81.64          & 85.97          & 69.91          & 77.95          & 79.91          & 86.03          & 79.63          & 85.81          \\
                                                             & Kimi K2.5 (no-think)~\cite{Kimi_K2_5_2026} & 2026.02 & 91.44          & 95.36          & 82.91          & 85.79          & 82.51          & 89.71          & 78.75          & 86.95          & 80.34          & 87.85          \\
                                                             & Gemini 2.5 Pro~\cite{2025gemini}           & 2025.03 & 90.90          & 94.32          & 85.56          & 90.07          & 88.94          & 89.47          & 79.46          & 87.13          & 81.66          & 88.08          \\
                \midrule
                \multirow{9}{*}{\makecell{Document \nextline Parsing \nextline VLMs}}
                                                             & MonkeyOCR-pro-1.2B~\cite{li2025monkeyocr}  & 2025.07 & 84.24          & 89.02          & 68.41          & 75.62          & 67.21          & 72.81          & 67.98          & 72.91          & 69.20          & 74.29          \\
                                                             & MonkeyOCR-pro-3B~\cite{li2025monkeyocr}    & 2025.07 & 86.78          & 90.63          & 71.42          & 77.07          & 76.81          & 79.69          & 72.26          & 77.04          & 73.85          & 78.39          \\
                                                             & DeepSeek-OCR~\cite{wei2025deepseek}        & 2025.10 & 83.79          & 87.86          & 68.95          & 75.22          & 82.64          & 87.33          & 68.70          & 76.84          & 71.39          & 78.76          \\
                                                             & POINTS-Reader~\cite{liu2025points}         & 2025.08 & 77.13          & 81.66          & 64.29          & 73.91          & 73.96          & 82.13          & 72.03          & 81.13          & 72.28          & 80.94          \\
                                                             & FD-RL~\cite{FD_RL_2026}                    & 2025.11 & 87.35          & 92.10          & 72.81          & 78.29          & 76.95          & 82.39          & 74.31          & 80.51          & 75.55          & 81.52          \\
                                                             & dots.ocr~\cite{Dots_ocr_2025}              & 2025.07 & 88.62          & 92.86          & 75.42          & 81.65          & 82.04          & 86.27          & 73.39          & 81.84          & 75.61          & 83.17          \\
                                                             & PaddleOCR-VL~\cite{PaddleOCR_VL_2025}      & 2025.10 & 91.12          & 94.62          & 79.62          & 85.04          & 79.29          & 83.93          & 77.55          & 84.08          & 78.90          & 84.94          \\
                                                             & MinerU2.5~\cite{Mineru_2_5_2025}           & 2025.09 & 90.85          & 94.68          & 79.76          & 85.16          & 87.13          & 90.62          & 77.41          & 84.31          & 79.62          & 85.84          \\
                                                             & HunyuanOCR~\cite{HunyuanOCR_2025}          & 2025.11 & 95.74          & 97.71          & 81.44          & 86.91          & 82.34          & 87.74          & 77.55          & 85.22          & 79.67          & 86.55          \\
                \midrule
                \multirow{2}{*}{\textbf{Ours}}               & \textbf{StrucTab-SFT}                      & 2026.06 & 94.62          & 96.51          & 84.23          & 89.91          & 89.33          & 91.69          & 84.59          & 90.02          & 85.87          & 90.70          \\
                                                             & \textbf{StrucTab}                          & 2026.06 & \textbf{95.95} & \textbf{97.89} & \textbf{86.94} & \textbf{91.75} & \textbf{91.39} & \textbf{94.40} & \textbf{87.81} & \textbf{92.74} & \textbf{88.79} & \textbf{93.28} \\
                \bottomrule
            \end{tabular}
        }
    }
    \vspace{-1mm}
    \label{tab:experiment_results}
\end{table}

%% file: table/SFT_ablation.tex
\begin{table}[t!]
    \centering
    \captionsetup{font={small}}
    \caption{
        \textbf{Impact of Structural Decomposition via Sequential Curriculum.} Results on three open-source benchmarks (averaged) and the proposed TableVerse-5K.
    }
    \vspace{-0.4em}
    {\renewcommand{\arraystretch}{0.95}
        \setlength{\tabcolsep}{6pt}
        \resizebox{\columnwidth}{!} {
            \begin{tabular}{l cccccccc}
                \toprule
                \multirow{2}{*}[-2.5mm]
                {\textbf{Setting}}          &
                \multicolumn{3}{c}
                {\textbf{Task Design}}      &
                \multicolumn{2}{c}
                {\textbf{Open-Source Avg.}} &
                \multicolumn{2}{c}
                {\textbf{TableVerse-5K}}
                \\
                \cmidrule(lr){2-4}
                \cmidrule(lr){5-6}
                \cmidrule(lr){7-8}
                                            &
                \makecell{Row--column
                \\Counting} &
                \makecell{Merged-Cell
                \\Analysis} &
                \makecell{Sequential
                \\Reason} &
                TEDS                        &
                TEDS-S                      &
                TEDS                        &
                TEDS-S                                                                                                                     \\
                \midrule
                VLM backbone                & -      & -      & -      & 86.69          & 90.95          & 77.55          & 85.22          \\
                (a) Baseline                & -      & -      & -      & 86.26          & 90.37          & 81.12          & 85.51          \\
                (b) +RC                     & \cmark & -      & -      & 87.43          & 91.20          & 82.09          & 86.79          \\
                (c) +MC                     & -      & \cmark & -      & 87.39          & 91.28          & 82.39          & 87.01          \\
                (d) +RC +MC                 & \cmark & \cmark & -      & 88.93          & 91.94          & 83.24          & 88.61          \\
                (e) Seq (w/o Curr)          & -      & -      & \cmark & 89.54          & 92.41          & 84.12          & 89.40          \\
                \midrule
                (f) StrucTab-SFT            & \cmark & \cmark & \cmark & \textbf{90.11} & \textbf{92.97} & \textbf{84.59} & \textbf{90.02} \\
                \bottomrule
            \end{tabular}
        }
    }
    \vspace{-1mm}
    \label{tab:SFT_ablation}
\end{table}

%% file: table/Uni_TabRL.tex
\begin{table}[t!]
    \centering
    \captionsetup{font={small}}
    \caption{
        \textbf{Effectiveness of Uni-TabRL}.
        Both the Anchor-Guided Destylization mechanism and the 1D Probe structural reward contribute to overall performance gains, with their combination achieving the best overall results.
    }
    \vspace{-0.6em}
    {\renewcommand{\arraystretch}{0.95}
        \setlength{\tabcolsep}{6pt}
        \resizebox{\columnwidth}{!} {%
            \begin{tabular}{l cccccccc}
                \toprule
                \multirow{2}{*}[-0.4mm]
                {\textbf{Setting}}          &
                \multicolumn{3}{c}
                {\textbf{RL reward design}} &
                \multicolumn{2}{c}
                {\textbf{Open-Source Avg.}} &
                \multicolumn{2}{c}
                {\textbf{TableVerse-5K}}
                \\
                \cmidrule(lr){2-4}
                \cmidrule(lr){5-6}
                \cmidrule(lr){7-8}
                                            & Validity Reward & Structure Reward & Content Reward              & TEDS           & TEDS-S         & TEDS           & TEDS-S         \\
                \midrule
                StrucTab-SFT                & -               & -                & -                           & 90.11          & 92.97          & 84.59          & 90.02          \\
                (a) RL baseline             & \cmark          & TEDS-S           & TEDS                        & 91.29          & 94.02          & 86.45          & 91.67          \\
                (b) +VLM judge              & \cmark          & TEDS-S           & VLM-based consistency judge & 90.51          & 93.44          & 85.89          & 91.57          \\
                (c) +Anchor                 & \cmark          & TEDS-S           & Anchor-Guided Destylization & 91.76          & 94.70          & 87.10          & 92.16          \\
                (d) +1D Probe               & \cmark          & 1D Probe         & TEDS                        & 91.80          & 94.68          & 87.07          & 92.06          \\
                \midrule
                (e) \textbf{StrucTab}       & \cmark          & 1D Probe         & Anchor-Guided Destylization & \textbf{92.05} & \textbf{95.06} & \textbf{87.81} & \textbf{92.74} \\
                \bottomrule
            \end{tabular}
        }
    }
    \vspace{-3mm}
    \label{tab:Uni_TabRL}
\end{table}

%% file: table/RL_on_repetition.tex
\begin{table}[t!]
    \centering
    \captionsetup{font={small}}
    \caption{
        \textbf{1D Probe Structural Reward Reduces Repetition.}
        Compared with TEDS-S, the proposed 1D Probe structural reward leads to fewer repetitive generations.
    }
    {
        \renewcommand{\arraystretch}{0.92}
        \setlength{\tabcolsep}{8pt}
        \resizebox{\columnwidth}{!} {
            \begin{tabular}{ll cc}
                \toprule
                \multirow{2}{*}[-0.4mm]
                {\textbf{Setting}}          &
                \multirow{2}{*}[-0.4mm]
                {\textbf{RL reward design}} &
                \multicolumn{2}{c}
                {\textbf{TableVerse-5K}}                                                                          \\
                \cmidrule(lr){3-4}
                                            &                                   & \# Repetition & Repetition Rate \\
                \midrule
                VLM backbone                & -                                 & 94            & 1.88\%          \\
                StrucTab-SFT                & -                                 & 73            & 1.46\%          \\
                RL                          & TEDS + TEDS-S                     & 60            & 1.20\%          \\
                RL                          & TEDS + 1D Probe structural reward & \textbf{51}   & \textbf{1.02\%} \\
                \bottomrule
            \end{tabular}
        }
    }
    \label{tab:RL_on_repetition}
\end{table}

%% file: figure_code/qualitative_result.tex
\begin{figure}[t!]
    \centering
    \includegraphics[width=\linewidth]{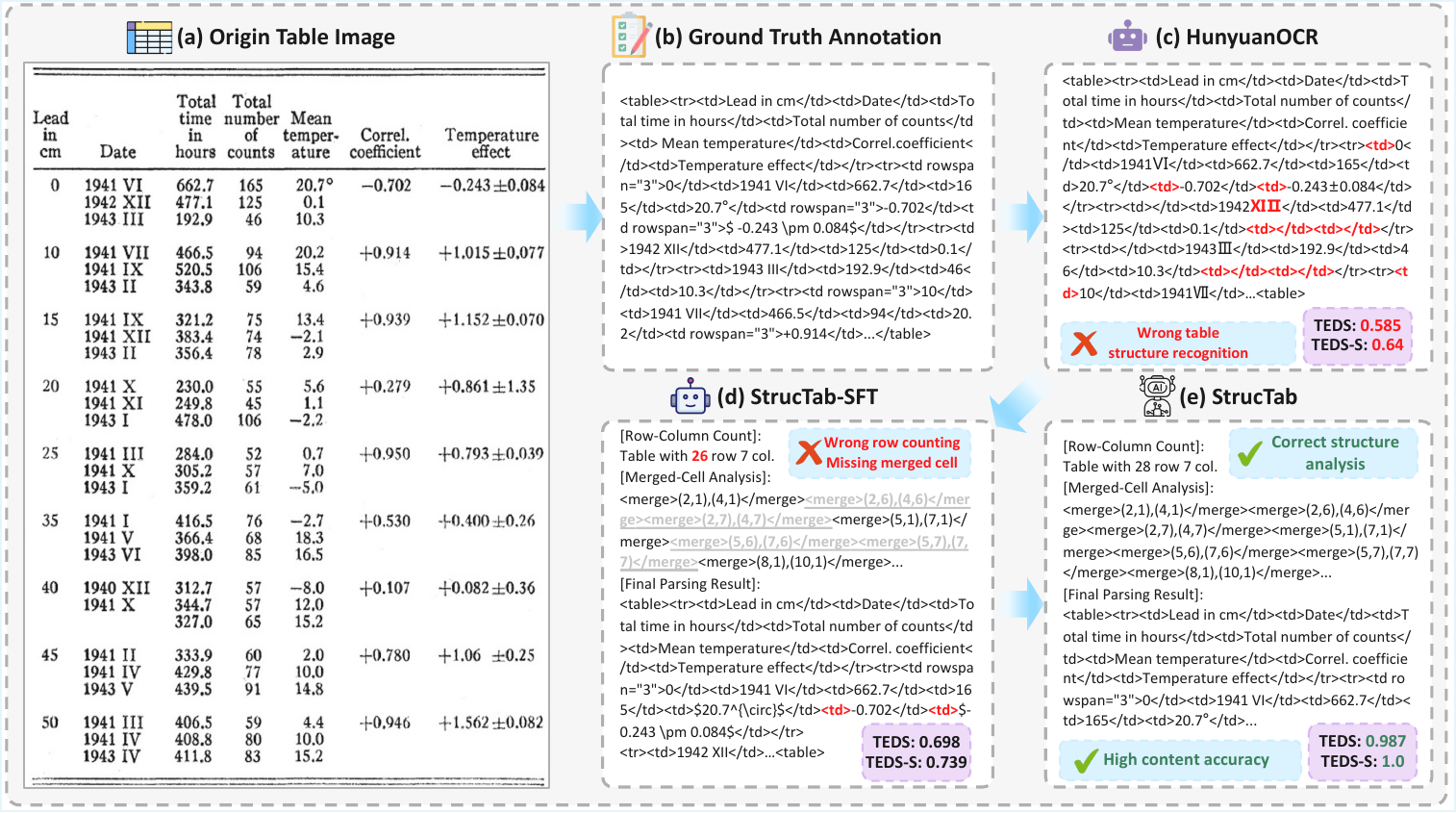}
    \vspace{-0.9em}
    \caption{
        \textbf{Qualitative Comparisons on TableVerse-5K.}
        Demonstrating progressive improvement: structural decomposition improves layout consistency in StrucTab-SFT, while Uni-TabRL further enhances structural alignment and content fidelity.
    }
    \label{fig:qualitative_result}
\end{figure}

%% file: section/6_conclusion.tex
\section{Concluding Remarks}
\label{sec:conclusion}

\mypara{Further Discussion.}
We further analyze why the early-error-focused 1D Probe structural reward is effective. Due to the autoregressive nature of VLMs, early tokens causally influence later predictions, so initial mistakes tend to propagate and amplify downstream errors~\cite{Hallucination_snowball_2023, Hallucination_snowball_2024}. Prior work shows that targeting the first failure point is particularly effective: Step-DPO~\cite{Step_DPO_2024} improves mathematical reasoning by optimizing the first incorrect step, and SENTINEL~\cite{SENTINEL_2025} reduces hallucination by intervening at the first erroneous sentence.
Consistently, our verification experiment in~\cref{subfig:error_rate_and_logp_comparison} shows that correcting the first structural error reduces subsequent structural errors by 75.25\% and increases generation confidence by 27.85\% (in logps). Motivated by this causal sensitivity, we introduce a 1D Probe structural reward that identifies the earliest structural deviation and provides localized RL feedback, improving structural reliability.

\mypara{Summary.}
In this paper, we propose StrucTab, a table parsing model learned through intermediate structural supervision and reward decomposition. At the modeling level, it decomposes parsing into human-inspired subtasks and progressively integrates them via sequential reasoning. To stabilize reinforcement learning, we introduce Uni-TabRL, which employs fine-grained decomposed rewards, including a 1D Probe for early structural feedback and an Anchor-Guided Destylization strategy for robust, style-invariant content supervision. Extensive experiments show state-of-the-art performance on public benchmarks and our challenging TableVerse-5K benchmark, validating the effectiveness of explicit structural modeling and reward decomposition.

%% file: section/7_appendix.tex
\appendix
\resetlinenumber[1]
\counterwithin{figure}{section}
\counterwithin{table}{section}
\maketitlesupplementary
\setcounter{page}{1}

\section*{Overview}
This material provides supplementary details to the main paper, including the following sections:
\vspace{-0.5em}
\begin{itemize}[
        label=\raisebox{0.5ex}{\tiny$\bullet$},
        leftmargin=1em,
        itemsep=0pt, % 项与项之间的间距
        parsep=2pt, % 段落之间的间距
        % topsep=0pt, % 列表与上下文的间距
        partopsep=0pt % 在新段落开始时的额外间距
    ]
    \item (\ref{sec:supp_method}) \textbf{Method Details}
          \begin{itemize}[label=-]
              \item (\ref{subsec:supp_decomposition}) Structural Decomposition Details
              \item (\ref{subsec:supp_Uni_TabRL}) Uni-TabRL Details
          \end{itemize}
    \item (\ref{sec:supp_tableverse}) \textbf{TableVerse-5K Details}
    \item (\ref{sec:supp_training}) \textbf{Training Details}
          \begin{itemize}[label=-]
              \item (\ref{subsec:supp_training_setup}) Training Setup
              \item (\ref{subsec:supp_training_data}) Training Data
              \item (\ref{subsec:supp_training_dynamics}) RL Training Dynamics
          \end{itemize}
    \item (\ref{sec:supp_evaluation}) \textbf{Evaluation Details}
          \begin{itemize}[label=-]
              \item (\ref{subsec:supp_evaluation_benchmarks}) Evaluation Benchmarks
              \item (\ref{subsec:supp_evaluation_counterparts}) Evaluation Counterparts
              \item (\ref{subsec:supp_evaluation_setup}) Evaluation Setup
          \end{itemize}
    \item (\ref{sec:supp_further_experiments}) \textbf{Further Experiments}
          \begin{itemize}[label=-]
              \item (\ref{subsec:supp_data_scale}) Effect of Training Data Scale
              \item (\ref{subsec:supp_backbone}) Generalization across VLM Backbones
              \item (\ref{subsec:supp_curriculum_necessity}) Necessity of Curriculum Training
              \item (\ref{subsec:supp_hard_samples}) Fine-grained Analysis on Hard Samples
          \end{itemize}
    \item (\ref{sec:supp_discussions}) \textbf{Discussions}
          \begin{itemize}[label=-]
              \item (\ref{subsec:supp_early_error}) Early Error Correction Verification
              \item (\ref{subsec:supp_destylization}) Anchor-Guided Destylization
              \item (\ref{subsec:supp_RL_efficiency}) Comparative Analysis of RL Efficiency
              \item (\ref{subsec:inference_efficiency}) Inference Efficiency of Sequential Reasoning
          \end{itemize}
    \item (\ref{sec:supp_limitation}) \textbf{Limitation and Future Work}
\end{itemize}

\newpage

\section{Method Details}
\label{sec:supp_method}

This section provides additional implementation details of our method. We first describe the structural decomposition and the corresponding annotation pipeline in~\cref{subsec:supp_decomposition}. We then present the detailed design and implementation of Uni-TabRL in~\cref{subsec:supp_Uni_TabRL}, including the 1D Probe structural reward, the HTML-to-image rendering pipeline, and the overall RL reward formulation.

\subsection{Structural Decomposition Details}
\label{subsec:supp_decomposition}

In this section, we provide details of the proposed structural decomposition. The annotations for the two subtasks (row--column counting and merged-cell analysis) are directly derived from standard HTML table annotations. Both subtasks are lightweight and enable scalable supervision without additional manual labeling. The construction pipeline follows the procedure described in~\cref{alg:data_pipeline}.

\input{algorithm/data_pipeline}

\subsection{Uni-TabRL Details}
\label{subsec:supp_Uni_TabRL}

In this section, we provide implementation details of Uni-TabRL.

\mypara{1D Probe Structural Reward.}
The core idea of this design is to linearize the two-dimensional table structure into a one-dimensional sequence for sequential comparison. We first use the Beautiful Soup Python library to remove all textual content from the `td' and `th' cells in both the predicted and reference tables, retaining only structural information (tag names, attributes, and row--column relations). The cells are then extracted in a row-wise, left-to-right order together with their row indices, forming two one-dimensional cell sequences. Starting from the beginning, the cells are compared pairwise by checking three conditions: a) whether they share the same row index, b) whether the tag names match, and c) whether the attribute dictionaries are identical. The comparison stops immediately once the first mismatch occurs. The final score is computed as the ratio between the number of successfully matched cells and the total number of cells in the reference table, yielding a value in \([0,1]\). This score measures the structural prefix alignment between the predicted and reference tables. The full computation procedure is illustrated in~\cref{alg:1D_probe_structural_reward}.

\mypara{Render table HTML to Image.}
This process is fully automated. We use Microsoft Playwright to control the Chromium engine for rendering HTML table pages. Mathematical formulas are rendered using MathJax, and Chinese text is displayed with the Noto Sans CJK font. The rendering pipeline is wrapped as a FastAPI service to facilitate efficient calls during RL training. This standardized pipeline consistently converts HTML tables into clear and easily parsable table images. The full rendering workflow is shown in~\cref{alg:render_HTML_to_PNG}.

To guarantee that the rendered images are visually standardized, the rendering template fixes all stylistic factors that are irrelevant to table structure and content. Specifically, we adopt a unified style sheet with a collapsed border model, a 2px solid outer border and 1px solid inner cell borders, a fixed cell padding of 6px vertically and 12px horizontally, and a plain white background. Empty \texttt{<td>} and \texttt{<th>} cells are filled with a non-breaking space placeholder to prevent zero-height rows. Mathematical formulas are typeset with the MathJax \texttt{tex-svg} backend (with \texttt{inlineMath} delimited by \texttt{\$} and \texttt{displayMath} by \texttt{\$\$}), and the page is rendered at a device scale factor of 2 within a $2000\times2000$ viewport. To ensure complete and stable rendering, we wait for both the MathJax startup promise and the browser font-ready signal, and invoke MathJax typesetting twice before capturing the screenshot. Finally, the image is tightly cropped to the bounding box of the rendered table with a small constant padding, removing redundant whitespace. These fixed conventions ensure that visually identical tables are always rendered into consistent images, which is essential for the Anchor-Guided Destylization reward.

\mypara{Overall RL Reward.}
The overall reward consists of three components: validity reward, structural reward, and content reward. Together, they provide comprehensive and fine-grained feedback for the table parsing task. The detailed workflow is described in~\cref{alg:RL_reward}.

\input{algorithm/1D_probe_structural_reward}
\input{algorithm/render_HTML_to_PNG}
\input{algorithm/RL_reward}

\input{section/8_appendix/8_2_tableverse}

\section{Training Details}
\label{sec:supp_training}

In this section, we provide additional training details, aiming to present the experimental setup more comprehensively and facilitate reproducibility.

\subsection{Training Setup}
\label{subsec:supp_training_setup}

We provide the detailed training configurations used in different stages.

\mypara{Pretraining \& SFT.}
We summarize the detailed training configurations for the pretraining and supervised fine-tuning (SFT) stages in~\cref{tab:training_setup_SFT}. Full-parameter fine-tuning is applied in all stages with a maximum context length of 8192 tokens and a maximum image resolution of 2K.

\mypara{RL.}
The RL configuration is summarized in~\cref{tab:training_setup_RL}. We train on 30K samples for 3 epochs with a batch size of 128, which takes approximately one day. To encourage exploration during training, we disable the KL loss and use a temperature of 1.0 for sampling, allowing the model to generate more diverse responses. We adopt a constant learning rate of $1\times10^{-6}$, which shows stable training behavior in our preliminary experiments. For the decomposed reward, we use fixed weights of $\lambda_1=0.4$, $\lambda_2=0.3$, and $\lambda_3=0.3$ throughout training.

\input{table/training_setup_SFT}
\input{table/training_setup_RL}

\subsection{Training Data}
\label{subsec:supp_training_data}

\mypara{Large-scale Synthetic Training Data Construction.}
To construct large-scale synthetic training data, we develop an automatic table generation pipeline based on LLMs. Our pipeline consists of three stages: table structure generation, structure normalization, and content population with visual rendering.

\textbf{Table structure generation.}
We first prompt an LLM to generate diverse HTML table structures.
The prompt instructs the model to produce valid table layouts with varying numbers of rows, columns, and merged-cell configurations.
An example prompt used for generating table structures is shown in~\cref{tab:synthetic_prompt}.
The generated HTML tables are then subjected to strict rule-based filtering to remove invalid structures, including cases where rows contain inconsistent column numbers or where merged-cell definitions introduce structural conflicts.

\textbf{Structure normalization.}
After filtering, we perform structural normalization on the remaining HTML tables to remove stylistic noise.
Specifically, only the core structural tags (\texttt{<table>}, \texttt{<tr>}, and \texttt{<td>}) and merge-related attributes (\texttt{rowspan} and \texttt{colspan}) are retained, while all styling-related elements (e.g., \texttt{style}, inline CSS, and formatting attributes) are removed.
This process produces a curated set of approximately 20K clean table structure templates that cover diverse structural patterns.

\textbf{Content population and rendering.}
Finally, we prompt the LLM again to populate each table template with diverse cell content, including natural language text and mathematical expressions.
The resulting HTML tables are then rendered into images using a browser-based rendering engine.
During rendering, we introduce additional visual variations by randomly assigning background colors to cells and sampling different font styles, including both printed and handwritten fonts.
This pipeline enables efficient generation of large-scale table-image pairs with diverse layouts, visual styles, and content distributions for pretraining.

\input{table/synthetic_prompt}

\mypara{Task Formulation Across Training Stages.}
In addition to the data construction pipeline, we further specify task formulations in different training stages.

\textbf{Pretrain-Stage1.}
In the first pretraining stage, we formulate table parsing as three independent subtasks, each with its own prompt and target format.

\emph{Row--column counting.}
The prompt is \emph{Count the number of rows and columns.}
The target output follows the format
\[
    \text{Table with } x \text{ rows, } y \text{ cols.}
\]
where $x$ and $y$ denote the row and column counts of the table, respectively.

\emph{Merged-cell analysis.}
The prompt is \emph{Identify all merged cells.}
Merged regions are represented using a structured token format
\[
    \langle \text{merge}\rangle(r_1,c_1),(r_2,c_2) \langle \text{/merge}\rangle ,
\]
which denotes a merged region spanning from row $r_1$, column $c_1$ to row $r_2$, column $c_2$.
When multiple merged regions exist, they are concatenated in a fixed order: rows are traversed from top to bottom, and within the same row, columns are traversed from left to right.

\emph{HTML parsing.}
For the standard end-to-end parsing task, the prompt is \emph{Parse this table into HTML format.}
The target output is the normalized HTML representation of the input table.

\textbf{Pretrain-Stage2, SFT, and RL.}
In the second pretraining stage and all subsequent stages (SFT and RL), we adopt the proposed curriculum training with sequential reasoning to unify the three subtasks into a single-round dialogue format. We emphasize that the SFT stage does not train the three subtasks separately; instead, it operates entirely on the unified sequential-reasoning format described below.
The prompt is \emph{Parse the table step by step.}
The model generates outputs sequentially, following the reasoning order
\[
    \text{row--column counting} \rightarrow
    \text{merged-cell analysis} \rightarrow
    \text{HTML parsing}.
\]
This formulation explicitly enforces a structured reasoning trajectory, allowing the model to first infer the global layout, then resolve local structural dependencies, and finally produce the complete HTML table representation.

\subsection{RL Training Dynamics}
\label{subsec:supp_training_dynamics}

To further illustrate the training behavior, we present the RL training dynamics in~\cref{fig:rl_dynamics}. As shown in the figure, the proposed Uni-TabRL framework supports stable reinforcement learning and provides a consistently improving reward signal throughout training. The (a) average reward steadily increases as the policy is optimized, indicating that the model progressively learns to produce structurally and semantically better table predictions.

In addition to the reward trajectory, other training statistics also demonstrate stable optimization behavior. The (c) average response length remains relatively consistent during training, suggesting that the model does not resort to degenerate strategies such as excessively short or repeated outputs. Meanwhile, the (d) gradient norm stays within a stable range without sudden spikes, indicating well-controlled optimization dynamics. Together, these observations confirm that Uni-TabRL provides a reliable training signal and enables robust policy optimization for table parsing.

\input{figure_code/rl_dynamics}

\section{Evaluation Details}
\label{sec:supp_evaluation}
In this section, we provide additional details about the evaluations in our experiments. These details clarify the evaluation protocol and implementation settings, ensuring a fair and reproducible comparison across different models.

\subsection{Evaluation Benchmarks}
\label{subsec:supp_evaluation_benchmarks}

In this section, we provide a detailed description of the public benchmarks used in our evaluation.

\mypara{OCRBench v2.}
OCRBench v2~\cite{OCRBench_v2_2026} is a comprehensive OCR evaluation benchmark designed to assess the capabilities of vision-language models across diverse OCR-related tasks.
In our experiments, we adopt its table parsing subset, which contains 700 table images collected from real-world documents.
These samples cover a wide range of layouts and visual complexities, providing a challenging testbed for evaluating table parsing performance.

\mypara{CC-OCR.}
CC-OCR~\cite{CC_OCR_2025} is a benchmark focusing on complex document OCR scenarios.
The dataset contains 300 scanned and photographed table images, including challenging cases such as long tables, complicated structures, and handwritten tables.
These characteristics make CC-OCR particularly suitable for evaluating model robustness under real-world imaging conditions.

\mypara{OmniDocBench.}
OmniDocBench~\cite{Omnidocbench_2025} is a large-scale document understanding benchmark covering multiple document analysis tasks.
We adopt the table parsing subset from OmniDocBench v1.5, which contains 512 table samples.

\subsection{Evaluation Counterparts}
\label{subsec:supp_evaluation_counterparts}

To provide a comprehensive evaluation, we compare our method with a diverse set of representative approaches covering three major categories: (a) expert table parsing (TP) models specifically designed for table structure recognition, (b) general-purpose vision-language models (VLMs) capable of performing table parsing through unified multimodal reasoning, and (c) document parsing VLMs that aim to process complex documents containing multiple structural elements.
Below, we briefly introduce the models used in our comparison.

\mypara{Expert TP Models.}
Expert TP models are dedicated architectures specifically designed for table structure recognition and parsing. These methods typically incorporate specialized representations or training objectives for tabular layouts.

\begin{itemize}[
        label=\raisebox{0.5ex}{\tiny$\bullet$},
        leftmargin=1em,
        itemsep=0pt, % 项与项之间的间距
        parsep=2pt, % 段落之间的间距
        % topsep=0pt, % 列表与上下文的间距
        partopsep=0pt % 在新段落开始时的额外间距
    ]
    \item \textbf{UniTable.} UniTable~\cite{UniTable_2024} is a unified training framework for table recognition that integrates the training paradigm, model architecture, and objective into a single formulation. It combines the simplicity of pixel-level inputs with the scalability enabled by self-supervised pretraining (SSP) on large collections of unlabeled table images. UniTable unifies the three core table recognition tasks (table structure extraction, cell content recognition, and cell bounding box prediction) under a task-agnostic language modeling objective, allowing the model to jointly learn multiple parsing capabilities within a single framework.
    \item \textbf{TRivia.} TRivia~\cite{TRivia_2025} is a self-supervised fine-tuning method that enables pretrained VLMs to learn table recognition directly from unlabeled table images collected in the wild. Built upon GRPO, TRivia automatically identifies unlabeled samples that are most beneficial for learning and eliminates the need for manual annotations. It introduces a question--answer--based reward mechanism where an attention-guided module generates diverse questions for each table image. The model receives feedback based on its ability to interpret the recognition results and correctly answer the generated questions, forming a closed-loop training process that allows the model to autonomously improve its capability in recognizing, structuring, and reasoning over tables.
\end{itemize}

\mypara{General-Purpose VLMs.}
General-purpose VLMs are large multimodal models trained on large-scale image--text data and capable of performing a wide variety of multimodal tasks, including table parsing. In our evaluation, we test several leading proprietary and open-source models to provide a broad comparison.
The proprietary models include GPT-4o~\cite{GPT_4o_2024}, GPT-5~\cite{GPT_5_2025}, Seed-1.8~\cite{Seed_1_8_2025}, and Gemini 2.5 Pro~\cite{2025gemini}, while the open-source models include Qwen2.5-VL~\cite{Qwen2.5-VL}, InternVL3.5~\cite{InternVL3_5_2025}, Qwen3-VL~\cite{Qwen3-VL}, and Kimi K2.5~\cite{Kimi_K2_5_2026}.
These models are not specifically designed for table parsing but have demonstrated strong general multimodal reasoning and document understanding abilities, making them valuable references for evaluating the effectiveness of specialized approaches.

\mypara{Document Parsing VLMs.}
Document parsing VLMs are designed to process complex document layouts and jointly recognize multiple structural elements such as text, tables, formulas, and figures. These models typically integrate layout understanding with multimodal reasoning and have recently achieved strong performance across a wide range of document analysis benchmarks.

\begin{itemize}[
        label=\raisebox{0.5ex}{\tiny$\bullet$},
        leftmargin=1em,
        itemsep=0pt, % 项与项之间的间距
        parsep=2pt, % 段落之间的间距
        % topsep=0pt, % 列表与上下文的间距
        partopsep=0pt % 在新段落开始时的额外间距
    ]
    \item \textbf{MonkeyOCR.} MonkeyOCR~\cite{MonkeyOCR_v1_5_2025} adopts a Structure--Recognition--Relation (SRR) triplet paradigm to simplify traditional modular document processing pipelines. Instead of relying on multiple independent tools, the SRR framework jointly models structural understanding, element recognition, and relational reasoning within a unified architecture. This design reduces the complexity of multi-stage pipelines while avoiding the inefficiency of applying large multimodal models to entire document pages.
    \item \textbf{DeepSeek-OCR.} DeepSeek-OCR~\cite{wei2025deepseek} explores the feasibility of compressing long document contexts through optical two-dimensional mapping. The system consists of two main components: DeepEncoder and a DeepSeek3B-MoE-A570M decoder. DeepEncoder serves as the core visual encoder designed to maintain low activation costs under high-resolution inputs while achieving high compression ratios. This design ensures that the number of visual tokens remains manageable, enabling efficient long document understanding.
    \item \textbf{POINTS-Reader.} POINTS-Reader~\cite{liu2025points} introduces an automated framework for constructing high-quality document extraction datasets and models capable of handling diverse document formats and layouts. The framework consists of two stages. The first stage generates large-scale and diverse synthetic training data, enabling the model to learn a unified representation for extracting key document elements. The second stage introduces a self-improvement strategy that further adapts the model trained on synthetic data to real-world documents, improving robustness and generalization.
    \item \textbf{FD-RL.} FD-RL~\cite{FD_RL_2026} proposes format decoupled RL for improving structured generation in document parsing tasks. The method leverages entropy-based data filtering to identify format-intensive instances that benefit most from targeted optimization. It further introduces format-specific rewards tailored to different structural patterns, enabling format-level validation rather than token-level memorization and improving the reliability of structured outputs.
    \item \textbf{dots.ocr.} dots.ocr~\cite{Dots_ocr_2025} proposes a unified end-to-end vision-language model that jointly learns three core document understanding tasks within a single framework. This capability is enabled by a highly scalable data engine that synthesizes large multilingual training corpora. By jointly optimizing multiple tasks across diverse languages, layouts, and domains, dots.ocr demonstrates strong generalization in large-scale document parsing scenarios.
    \item \textbf{PaddleOCR-VL.} PaddleOCR-VL~\cite{PaddleOCR_VL_2025} is a resource-efficient vision-language model designed specifically for document parsing, which integrates a NaViT-style dynamic-resolution visual encoder with the ERNIE-4.5-0.3B language model. This design enables accurate recognition of diverse document elements such as text, tables, formulas, and charts while maintaining relatively low computational cost. The model also supports multilingual document parsing across more than one hundred languages.
    \item \textbf{MinerU2.5.} MinerU2.5~\cite{Mineru_2_5_2025} is a 1.2B-parameter document parsing VLM that achieves strong recognition accuracy with high computational efficiency. It adopts a coarse-to-fine two-stage parsing strategy that separates global layout analysis from local content recognition. In the first stage, the model performs layout analysis on downsampled images to efficiently identify structural elements. In the second stage, guided by the predicted layout, it performs fine-grained recognition on high-resolution image crops, preserving detailed information for dense text, complex formulas, and tables.
    \item \textbf{HunyuanOCR.} HunyuanOCR~\cite{HunyuanOCR_2025} is a lightweight open-source model designed for OCR tasks. The architecture combines a Native ViT encoder with a lightweight language model connected through an MLP adapter. Despite having only around one billion parameters, HunyuanOCR demonstrates competitive performance across perception tasks such as text spotting and document parsing, as well as semantic tasks including information extraction and image translation. The model further benefits from high-quality training data and reinforcement learning strategies that significantly improve performance.
\end{itemize}

\subsection{Evaluation Setup}
\label{subsec:supp_evaluation_setup}

For evaluation, we maintain strict variable control to ensure reproducibility and fair comparison across different models.
We employ \texttt{vLLM} as the default inference engine for all compatible models to provide consistent decoding behavior and efficient batch inference.
Following recent work~\cite{TRivia_2025}, for general-purpose VLMs we use a decoding temperature of 0.2 to balance parsing accuracy and repetition, together with a unified prompt design (shown in~\cref{tab:VLM_prompt}).
For document-parsing VLMs, we adopt their officially recommended table recognition prompts to preserve their intended inference configurations.

To further ensure evaluation consistency, we reproduce the reported results of recent work~\cite{TRivia_2025} on overlapping benchmarks.
For general-purpose VLMs, the reproduced scores differ from the reported values by less than 2 points in both TEDS and TEDS-S, while for document-parsing VLMs, the deviation is within 1.5 points.
These small discrepancies indicate that our evaluation pipeline closely matches the original experimental settings.
For newly introduced baselines, we strictly follow the same inference protocol and prompt configuration to guarantee a fair and consistent evaluation across all compared models.

\begin{table}[t!]
    \centering
    \captionsetup{font={small}}
    \caption{\textbf{Unified Prompt for General-purpose VLMs.}}
    \vspace{-0.4em}
    \setlength{\tabcolsep}{10pt}
    \renewcommand{\arraystretch}{0.95}
    \begin{tabular}{p{0.9\columnwidth}}
        \toprule
        \textbf{Prompt}
        \\
        \midrule
        \texttt{<image>} You are an AI specialized in recognizing and extracting tables from images.
        Your mission is to analyze the table image and generate the result in HTML format using specified tags.
        Output only the results without any other words or explanations.
        \\
        \bottomrule
    \end{tabular}
    \vspace{-2mm}
    \label{tab:VLM_prompt}
\end{table}

\input{section/8_appendix/8_5_further_experiments}
\input{section/8_appendix/8_6_discussions}

\section{Limitation and Future Work}
\label{sec:supp_limitation}

While StrucTab significantly advances the robustness of table parsing, our current framework primarily focuses on single-page table images. Processing extremely long or multi-page tables remains challenging due to the inherent resolution and context-window limitations of current vision-language models. In future work, we plan to extend our structured reasoning paradigm to document-level scenarios, enabling cross-page table understanding and reconstruction.

Another limitation lies in handling highly irregular tables whose structural boundaries do not follow consistent row--column alignments, such as tables with misaligned separators, partially missing borders, or complex layout artifacts. These cases remain difficult for current models because the conventional HTML representation assumes a regular grid structure. For such irregular layouts, forcing the prediction into a strictly normalized HTML format may not always provide the most faithful representation of the original table structure.

In future work, we aim to explore more flexible structural representations that can better capture diverse table organizations beyond rigid grid assumptions. Investigating alternative structural abstractions and more expressive output formats may enable more robust modeling of real-world tables with complex or irregular layouts.

%% file: algorithm/data_pipeline.tex
\begin{algorithm}
    \caption{Construction of Sequential Reasoning Data for Table Parsing}
    \begin{algorithmic}[1]
        \Require raw Dataset $\mathcal{D}_{\text{raw}} = \{(I_i, H_i)\}_{i=1}^N$ containing table images and HTML annotations
        \Ensure Processed Dataset $\mathcal{D}_{\text{cot}}$ with reasoning chains
        \Statex

        \Function{ConstructCurriculumData}{$\mathcal{D}_{\text{raw}}$}
        \State $\mathcal{D}_{\text{cot}} \gets \emptyset$

        \For{\textbf{each} $(\text{image:}\; I,\text{HTML:}\; H) \in \mathcal{D}_{\text{raw}}$}
        \State \textbf{\# [Step 1] Subtask 1 - Row-Column Counting}
        \State $N_{\text{row}}, N_{col} \gets \text{ParseTableDimensions}(H)$
        \State $S_{\text{dim}} \gets \text{``}N_{\text{row}}\text{ rows and }N_{col}\text{ columns''}$

        \State
        \State \textbf{\# [Step 2] Subtask 2 - Merged-Cell Analysis}
        \State $H_{\text{empty}} \gets \text{StripTextContent}(H)$ \Comment{Retain only structure (tags \& attrs)}
        \State $S_{\text{merge}} \gets \text{ExtractMergeTokens}(H_{\text{empty}})$

        \State
        \State \textbf{\# [Step 3] Sequential Reasoning Data Construction}
        \State $Seq_{\text{reasoning}} \gets \text{``[Row-Column Analysis]: \textbackslash n''} \oplus S_{\text{dim}} \oplus \text{``\textbackslash n\textbackslash n''}$
        \State $Seq_{\text{reasoning}} \gets Seq_{\text{reasoning}} \oplus \text{``[Merged-Cell Analysis]: \textbackslash n''} \oplus S_{\text{merge}} \oplus \text{``\textbackslash n\textbackslash n''}$
        \State $Seq_{\text{reasoning}} \gets Seq_{\text{reasoning}} \oplus \text{``[Final Table Parsing]: \textbackslash n''} \oplus H$

        \State
        \State $\mathcal{D}_{\text{cot}} \gets \mathcal{D}_{\text{cot}} \cup \{ (I,  H, Seq_{\text{reasoning}}) \}$
        \EndFor

        \State \Return $\mathcal{D}_{\text{cot}}$
        \EndFunction

        \Statex
        \Function{ExtractMergeTokens}{$H_{\text{empty}}$}
        \State $Tokens \gets []$
        \State $Grid \gets \text{InitializeVirtualGrid}()$ \Comment{Track cell occupancy}

        \For{\textbf{each} $row \in H_{\text{empty}}$}
        \For{\textbf{each} $cell \in row$}
        \State $(r_1, c_1) \gets \text{FindNextEmptyPosition}(Grid)$
        \State $R_{span} \gets cell.\text{getAttribute(``rowspan'', 1)}$
        \State $C_{span} \gets cell.\text{getAttribute(``colspan'', 1)}$

        \State $(r_2, c_2) \gets (r_1 + R_{span} - 1, c_1 + C_{span} - 1)$
        \State $\text{MarkOccupied}(Grid, r_1, c_1, r_2, c_2)$

        \If{$R_{span} > 1$ \textbf{or} $C_{span} > 1$}
        \State $Token \gets \text{``\texttt{<merge>}(} r_1, c_1 \text{),(} r_2, c_2 \text{)\texttt{</merge>}''}$
        \State $Tokens.\text{Append}(Token)$
        \EndIf
        \EndFor
        \EndFor

        \If{$\text{Length}(Tokens) == 0$}
        \State \Return \text{``No merged cells found.''}
        \Else
        \State \Return $\text{Join}(Tokens, \text{`` ''})$
        \EndIf
        \EndFunction

    \end{algorithmic}
    \label{alg:data_pipeline}
\end{algorithm}

%% file: algorithm/1D_probe_structural_reward.tex
\begin{algorithm}
    \caption{1D Probe Structural Reward Calculation}
    \begin{algorithmic}[1]
        \Require Predicted HTML Table $H_{\text{pred}}$, Reference HTML Table $H_{\text{ref}}$
        \Ensure Structural Reward $R \in [0, 1]$
        \Statex
        \Function{Compute1DProbeReward}{$H_{\text{pred}}, H_{\text{ref}}$}
        \State \textbf{\# [Step 1] Structure Extraction}
        \State $S_{\text{pred}} \gets \text{CleanContent}(H_{\text{pred}})$ \Comment{Remove text, keep tags \& attrs}
        \State $S_{\text{ref}} \gets \text{CleanContent}(H_{\text{ref}})$
        \State
        \State \textbf{\# [Step 2] 1D Linearization}
        \State $L_{\text{pred}} \gets \text{ExtractCells}(S_{\text{pred}})$ \Comment{List of $(cell, row\_idx)$}
        \State $L_{\text{ref}} \gets \text{ExtractCells}(S_{\text{ref}})$
        \State $N_{\text{ref}} \gets \text{Length}(L_{\text{ref}})$
        \State $N_{\text{pred}} \gets \text{Length}(L_{\text{pred}})$
        \State
        \State \textbf{\# [Step 3] 1D Probing Reward Computation}
        \State $k \gets 0$ \Comment{Counter for matched cells}
        \State $Limit \gets \min(N_{\text{ref}}, N_{\text{pred}})$

        \For{$i \gets 0$ \textbf{to} $Limit - 1$}
        \State $(C_{p}, R_{p}) \gets L_{\text{pred}}[i]$ \Comment{Get cell object and row index}
        \State $(C_{r}, R_{r}) \gets L_{\text{ref}}[i]$
        \State \textbf{Check Mismatch Conditions:}
        \If{$R_{p} \neq R_{r}$} \textbf{break} \Comment{Row alignment error}
        \EndIf
        \If{$C_{p}.\text{tagName} \neq C_{r}.\text{tagName}$} \textbf{break} \Comment{Tag mismatch (td vs th)}
        \EndIf
        \If{$C_{p}.\text{attrs} \neq C_{r}.\text{attrs}$} \textbf{break} \Comment{Attr mismatch (colspan/rowspan)}
        \EndIf
        \State $k \gets k + 1$ \Comment{Cell is structurally correct}
        \EndFor
        \State
        \State $R \gets k / N_{\text{ref}}$
        \State \Return $R$
        \EndFunction
        \State
        \Function{ExtractCells}{$Soup$}
        \State $List \gets []$
        \State $Rows \gets Soup.\text{FindAll}(\text{``tr''})$
        \For{$r \gets 0$ \textbf{to} $\text{Length}(Rows) - 1$}
        \State $Cells \gets Rows[r].\text{FindAll}([\text{``td'', ``th''}])$
        \For{$c \in Cells$}
        \State $List.\text{Append}( (c, r) )$ \Comment{Store cell with its row index}
        \EndFor
        \EndFor
        \State \Return $List$
        \EndFunction
    \end{algorithmic}
    \label{alg:1D_probe_structural_reward}
\end{algorithm}

%% file: algorithm/render_HTML_to_PNG.tex
\begin{algorithm}
    \caption{HTML Table String to PNG Rendering Process}
    \begin{algorithmic}[1]
        \Require Input HTML string $S$, Output image name $N$, Constant configuration $C$
        \Ensure Rendered image file path $P_{\text{out}}$
        \Statex

        \Function{RenderHTMLToImage}{$S, N, C$}
        \State \textbf{\# [Step 1] Path Preparation}
        \State $N_{\text{safe}} \gets \text{SanitizeFileName}(N)$ \Comment{Prevent path traversal}
        \State $P_{\text{out}} \gets \text{AbsPath}(C.\text{OutDir}, N_{\text{safe}})$ \Comment{Output image path}
        \State
        \State \textbf{\# [Step 2] HTML Construction}
        \State $S' \gets \text{FillEmptyCells}(S)$ \Comment{Insert placeholder \&nbsp; for empty \texttt{td/th}}
        \State $H \gets \text{InjectMathJax}(S', C.\text{MathJaxURL})$ \Comment{Enable LaTeX Rendering}
        \State $H_{\text{full}} \gets \text{WrapInTemplate}(H)$ \Comment{Use constant HTML style}
        \State
        \State \textbf{\# [Step 3] Browser Interaction}
        \State $Page \gets BrowserContext.\text{NewPage}()$
        \State $Page.\text{SetContent}(H_{\text{full}})$
        \State \textbf{Wait} for \texttt{MathJax.startup.promise}
        \State \textbf{Wait} for \texttt{document.fonts.ready}
        \State $Page.\text{Evaluate}(\text{"MathJax.typesetPromise()"})$ \Comment{Final typeset}
        \State
        \State \textbf{\# [Step 4] Screenshot Area Calculation}
        \State $Target \gets Page.\text{Locate}([\text{Table}])$ \Comment{Get rendered table bbox}
        \State $Box \gets Target.\text{BoundingBox}()$
        \State $ClipRect \gets \text{AddPadding}(Box, C.\text{Padding})$
        \State $Page.\text{Screenshot}(path=P_{\text{out}}, clip=ClipRect)$
        \State $Page.\text{Close}()$ \Comment{Resource Cleanup}

        \State \Return $P_{\text{out}}$
        \EndFunction

    \end{algorithmic}
    \label{alg:render_HTML_to_PNG}
\end{algorithm}

%% file: algorithm/RL_reward.tex
\begin{algorithm}
    \caption{Decomposed RL Reward Calculation for Table Parsing}
    \begin{algorithmic}[1]
        \Require Predicted HTML $H_{\text{pred}}$, Reference HTML $H_{\text{ref}}$, Weights $[\lambda_{1}, \lambda_{2}, \lambda_{3}]$
        \Ensure Final Reward $R \in [0, 1]$
        \Statex
        \Function{ComputeReward}{$H_{\text{pred}}, H_{\text{ref}}$}
        \State \textbf{\# [Phase 1] Validity Reward: Hard Gating ($\lambda_{1}$)}
        \If{\textbf{not} $\text{HasValidTags}(H_{\text{pred}})$ \textbf{or} \textbf{not} $\text{HasTableEnd}(H_{\text{pred}})$}
        \State \Return $0.0$ \Comment{Invalid syntax yields zero total reward}
        \EndIf

        \State $Info_{\text{pred}} \gets \text{ExtractRowColInfo}(H_{\text{pred}})$
        \State $Info_{\text{ref}} \gets \text{ExtractRowColInfo}(H_{\text{ref}})$
        \State $Merge_{\text{pred}} \gets \text{ExtractMergeCells}(H_{\text{pred}})$
        \State $Merge_{\text{ref}} \gets \text{ExtractMergeCells}(H_{\text{ref}})$

        \If{$Info_{\text{pred}} \neq Info_{\text{ref}}$ \textbf{or} $Merge_{\text{pred}} \neq Merge_{\text{ref}}$}
        \State $R_{\text{valid}} \gets 0.0$ \Comment{Zero if any mismatch occurs}
        \Else
        \State $R_{\text{valid}} \gets 1.0$
        \EndIf
        \State
        \State \textbf{\# [Phase 2] Structure Reward: 1D Probe ($\lambda_{2}$)}
        \State $R_{\text{structure}} \gets \text{Compute1DProbeReward}(H_{\text{pred}}, H_{\text{ref}})$ \Comment{See~\cref{alg:1D_probe_structural_reward}}
        \State
        \State \textbf{\# [Phase 3] Content Reward: Anchor-Guided Destylization ($\lambda_{3}$)}
        \State $I^{\text{render}}_{\text{pred}} \gets \text{RenderHTMLToImage}(H_{\text{pred}})$ \Comment{See~\cref{alg:render_HTML_to_PNG}}
        \State $H^{\text{render}}_{\text{pred}} \gets \text{AnchorOCRModel}(I^{\text{render}}_{\text{pred}})$
        \State $I^{\text{render}}_{\text{ref}} \gets \text{RenderHTMLToImage}(H_{\text{ref}})$ \Comment{See~\cref{alg:render_HTML_to_PNG}}
        \State $H^{\text{render}}_{\text{ref}} \gets \text{AnchorOCRModel}(I^{\text{render}}_{\text{ref}})$
        \State $R_{\text{content}} \gets \text{TEDS}(H^{\text{render}}_{\text{pred}}, H^{\text{render}}_{\text{ref}})$
        \State
        \State \textbf{\# Weighted Aggregation}
        \State $R \gets \lambda_{1} \cdot R_{\text{valid}} + \lambda_{2} \cdot R_{\text{structure}} + \lambda_{3} \cdot R_{\text{content}}$
        \State \Return $R$
        \EndFunction
        \Statex
        \Statex \Comment{
            Since both the rendering process and the AnchorOCRModel are deterministic, the de-stylization of reference answers is pre-computed offline in our experiments.
        }
    \end{algorithmic}
    \label{alg:RL_reward}
\end{algorithm}

%% file: section/8_appendix/8_2_tableverse.tex
\section{TableVerse-5K Details}
\label{sec:supp_tableverse}

In this section, we provide additional details on the construction of the TableVerse-5K benchmark, focusing on the human annotation effort behind it.

\mypara{Human Annotation Protocol.}
Starting from the multi-VLM candidate drafts, each table image is refined by one of 20 expert annotators. On average, refining a single table draft takes about 6 minutes, with the most complex cases requiring up to 34 minutes, followed by roughly 4 minutes of validation. To guarantee annotation quality, every sample undergoes a rigorous three-round cross-verification process, and any instance that fails verification is discarded and re-annotated until it is defect-free. In addition, we perform strict deduplication between the training data and the benchmark to prevent leakage. We commit to open-sourcing the benchmark data to support community-wide research.

%% file: table/training_setup_SFT.tex
\begin{table*}[t]
    \centering
    \caption{\textbf{Training setup and hyperparameters for pretraining and SFT.}}
    \label{tab:training_setup_SFT}
    \vspace{-0.4em}
    \scriptsize
    \setlength{\tabcolsep}{6pt}
    \renewcommand{\arraystretch}{1.1}
    \begin{tabular}{c l >{\centering\arraybackslash}p{0.18\textwidth} >{\centering\arraybackslash}p{0.18\textwidth} >{\centering\arraybackslash}p{0.18\textwidth}}
        \toprule
         & \textbf{Setting}                     & \textbf{Stage-1}                                & \textbf{Stage-2}     & \textbf{SFT}         \\
        \midrule

        \multirow{2}{*}{\rotatebox[origin=c]{90}{\small\textit{ViT}}}
         & {Max Resolution}                     & 2K                                              & 2K                   & 2K                   \\
         & {Tokens per Image}                   & 256-1280                                        & 256-1280             & 256-1280             \\
        \midrule

        \multirow{3}{*}{\rotatebox[origin=c]{90}{\small\textit{Data}}}
         & {Dataset}                            & Synth + Public                                  & Synth + Public       & Public + Manual      \\
         & {Raw Samples}                        & 6M                                              & 6M                   & 100K                 \\
         & {Training Samples}                   & 18M                                             & 6M                   & 100K                 \\
        \midrule

        \multirow{4}{*}{\rotatebox[origin=c]{90}{\small\textit{Model}}}
         & {Trainable Parameters}               & All                                             & All                  & All                  \\
         & \multirow{2}{*}{{Training Paradigm}} & Independent Tasks                               & Sequential Reasoning & Sequential Reasoning \\
         & {Sequence Length}                    & 8192                                            & 8192                 & 8192                 \\
        \midrule

        \multirow{7}{*}{\rotatebox[origin=c]{90}{\small\textit{Training}}}
         & {Batch Size}                         & 256                                             & 128                  & 128                  \\
         & {Epoch}                              & 1                                               & 1                    & 2                    \\
         & {Learning Rate}                      & $4\times10^{-5}$                                & $4\times10^{-5}$     & $2\times10^{-5}$     \\
         & {min-LR}                             & $2\times10^{-6}$                                & $2\times10^{-6}$     & $1\times10^{-6}$     \\
         & {LR scheduler}                       & \multicolumn{3}{c}{Cosine}                                                                    \\

         & {Optimizer}                          & \multicolumn{3}{c}{Adam}                                                                      \\
         & {Hardware}                           & \multicolumn{3}{c}{16 $\times$ NVIDIA H20 GPUs}                                               \\
        \bottomrule
    \end{tabular}
\end{table*}

%% file: table/training_setup_RL.tex
\begin{table}[t!]
    \centering
    \captionsetup{font={small}}
    \caption{
        \textbf{Reinforcement Learning Training Configuration.}
    }
    \label{tab:training_setup_RL}
    \vspace{-0.4em}
    {\renewcommand{\arraystretch}{0.92}
        \setlength{\tabcolsep}{22pt}
        \resizebox{\columnwidth}{!}{
            \begin{tabular}{c l >{\centering\arraybackslash}p{0.28\textwidth}}
                \toprule
                \textbf{Group} & \textbf{Setting}         & \textbf{Value}        \\
                \midrule

                \multirow{4}{*}{{\small\textit{Model}}}
                               & Tensor Parallelism (TP)  & 1                     \\
                               & Micro batch size / GPU   & 1                     \\
                               & PPO mini-batch size      & 128                   \\
                               & ZeRO stage               & 3                     \\
                \midrule
                \multirow{5}{*}{{\small\textit{Optimization}}}
                               & Training batch size      & 128                   \\
                               & Optimizer                & Adam                  \\
                               & Learning rate            & $1\times10^{-6}$      \\
                               & LR schedule              & Constant              \\
                               & Warm-up steps            & 0                     \\
                \midrule
                \multirow{3}{*}{{\small\textit{Sequence}}}
                               & Sequence length          & 8192                  \\
                               & Max prompt length        & 4096                  \\
                               & Max response length      & 4096                  \\
                \midrule
                \multirow{5}{*}{{\small\textit{RL Algorithm}}}
                               & Advantage estimator      & GRPO                  \\
                               & GRPO clip ratio          & 0.2                   \\
                               & Use KL in reward         & False                 \\
                               & KL loss coefficient      & 0.0                   \\
                               & Entropy coefficient      & 0.0                   \\
                \midrule
                \multirow{4}{*}{{\small\textit{Rollout}}}
                               & Temperature              & 1.0                   \\
                               & Top-$p$                  & 0.95                  \\
                               & Top-$k$                  & 50                    \\
                               & Samples per prompt ($n$) & 8                     \\
                \midrule
                \multirow{2}{*}{{\small\textit{Data}}}
                               & Training data size       & 30K                   \\
                               & Total epochs             & 3                     \\
                \midrule
                \multirow{1}{*}{{\small\textit{Hardware}}}
                               & GPUs                     & $16\times$ NVIDIA H20 \\

                \bottomrule
            \end{tabular}
        }
    }
    \vspace{-2mm}
\end{table}

%% file: table/synthetic_prompt.tex
\begin{table}[t!]
    \centering
    \captionsetup{font={small}}
    \caption{\textbf{Prompt for Synthetic Table Structure Generation.}}
    \vspace{-0.4em}
    \setlength{\tabcolsep}{10pt}
    \renewcommand{\arraystretch}{0.95}
    \begin{tabular}{p{0.9\columnwidth}}
        \toprule
        \textbf{Prompt} \\
        \midrule
        \texttt{You are an AI assistant specialized in generating HTML tables.
            Your task is to produce valid HTML table structures with diverse layouts.
            The table should use only <table>, <tr>, and <td> tags, and may include
            merged cells using rowspan and colspan attributes.
            Ensure that the table structure is valid and consistent.
            Do not include any styling information such as CSS or style attributes.
            Output only the HTML table.
        }               \\
        \bottomrule
    \end{tabular}
    \vspace{-2mm}
    \label{tab:synthetic_prompt}
\end{table}

%% file: figure_code/rl_dynamics.tex
\begin{figure}[t!]
    \centering
    \includegraphics[width=\linewidth]{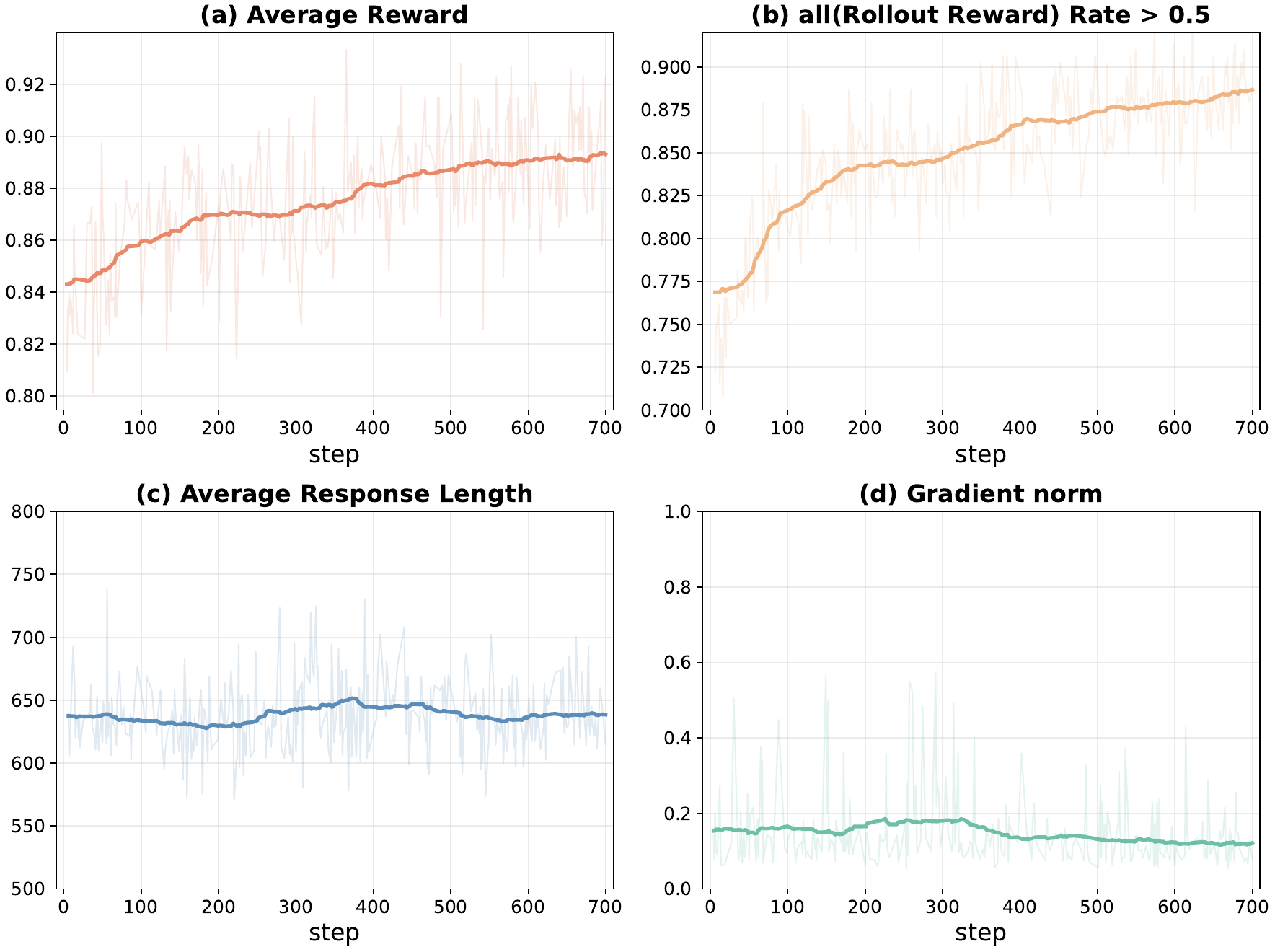}
    \vspace{-1.8em}
    \caption{
        \textbf{Uni-TabRL Training Dynamics.}
    }
    \vspace{-0.5em}
    \label{fig:rl_dynamics}
\end{figure}

%% file: section/8_appendix/8_5_further_experiments.tex
\section{Further Experiments}
\label{sec:supp_further_experiments}

In this section, we provide a set of supplementary experiments that further examine the source of StrucTab's improvements, its generality across backbones, the necessity of the decomposed curriculum, and its behavior on structurally challenging samples. Unless otherwise specified, all experiments are conducted under the same training configuration and environment as the main HunyuanOCR-based model described in~\cref{sec:supp_training}, ensuring a fair and reproducible comparison.

\subsection{Effect of Training Data Scale}
\label{subsec:supp_data_scale}

A natural concern is whether the gains of StrucTab mainly stem from the scale of pretraining data rather than from structural reasoning. To examine this, we vary the pretraining scale among 1M, 2M, and 6M samples while keeping the SFT stage identical to our main setting. As reported in Exp1 of~\cref{tab:further_experiments}, performance saturates between 2M and 6M: the 2M model is already comparable to the 6M model across both the open-source average and TableVerse-5K. This indicates that our pipeline effectively exploits the available data and is not heavily dependent on excessively large data scales.

\subsection{Generalization across VLM Backbones}
\label{subsec:supp_backbone}

To verify that the StrucTab paradigm is not tied to a specific backbone, we additionally instantiate it on InternVL3.5-1B~\cite{InternVL3_5_2025}. For a fair and reproducible comparison, the training configuration and environment are kept consistent with the HunyuanOCR-based model. As shown in Exp2 of~\cref{tab:further_experiments}, the StrucTab paradigm consistently outperforms the direct image-to-HTML baseline (6M\textsubscript{base}), and InternVL3.5-1B exhibits scaling trends similar to HunyuanOCR. These results confirm that the effectiveness of StrucTab is backbone-agnostic and transfers across different model families.

\subsection{Necessity of Curriculum Training}
\label{subsec:supp_curriculum_necessity}

Under the 6M setting, the decomposed curriculum yields only modest gains, which may suggest that it is redundant. We hypothesize that heavy-duty pretraining implicitly equips the model with single-shot reasoning ability, thereby masking the contribution of the curriculum. To test this, we remove the decomposed curriculum and train directly with sequential reasoning under different data scales. As shown in Exp3 of~\cref{tab:further_experiments}, removing the curriculum leads to consistent degradation, and the gap grows rapidly as data becomes scarcer, reaching up to roughly 4 points in TEDS-S at the 1M scale. This shows that the decomposed curriculum acts as a catalyst for structural awareness and becomes increasingly critical as data availability decreases.

\input{table/further_experiments}
\input{table/hard_sample_analysis}

\subsection{Fine-grained Analysis on Hard Samples}
\label{subsec:supp_hard_samples}

Aggregate benchmark scores may understate the benefit of our reward design, since easy samples are already well solved after SFT. To provide a finer-grained view, we isolate the hard-sample subset of TableVerse-5K, defined as the 1{,}021 samples whose TEDS-S is below 0.75 after SFT. These samples are structurally more complex than the full set, with an average of 19.31 rows and 9.52 columns versus 14.39 rows and 7.13 columns. As reported in~\cref{tab:hard_sample_analysis}, our RL scheme improves the SFT baseline from 0.415/0.514 to 0.572/0.682 (TEDS/TEDS-S), clearly surpassing the vanilla RL baseline at 0.524/0.623. The margin over vanilla RL is substantially larger on this hard subset than on the full benchmark, indicating that the benefit of our reward design is concentrated on structurally challenging cases and justifying its additional design complexity.

%% file: table/further_experiments.tex
\providecommand{\down}[1]{\textcolor{red}{\footnotesize(#1)}}
\begin{table}[t!]
    \centering
    \captionsetup{font={small}}
    \caption{
        \textbf{Further experiments on data scale, backbone generality, and curriculum necessity.}
        Exp1 studies the effect of the pretraining data scale; Exp2 verifies that the StrucTab paradigm is backbone-agnostic on InternVL3.5-1B; and Exp3 measures the degradation caused by removing the decomposed curriculum under different data scales. ``Curr.'' denotes whether the decomposed curriculum is used.
    }
    \vspace{-0.6em}
    {\renewcommand{\arraystretch}{0.95}
        \setlength{\tabcolsep}{10pt}
        \resizebox{\columnwidth}{!}{
            \begin{tabular}{cccccc}
                \toprule
                \multirow{2}{*}[-0.4mm]{\textbf{Scale}}       &
                \multirow{2}{*}[-0.4mm]{\textbf{Curr.}}       &
                \multicolumn{2}{c}{\textbf{Open-Source Avg.}} &
                \multicolumn{2}{c}{\textbf{TableVerse-5K}}
                \\
                \cmidrule(lr){3-4}
                \cmidrule(lr){5-6}
                                                              &        & TEDS                   & TEDS-S                 & TEDS                   & TEDS-S                 \\
                \midrule
                \multicolumn{6}{l}{\textit{Exp1: Scale of training data (HunyuanOCR)}}                                                                                     \\
                6M                                            & \cmark & 90.11                  & 92.97                  & 84.59                  & 90.02                  \\
                2M                                            & \cmark & 89.74                  & 92.88                  & 84.37                  & 90.00                  \\
                1M                                            & \cmark & 87.21                  & 90.47                  & 82.12                  & 87.73                  \\
                \midrule
                \multicolumn{6}{l}{\textit{Exp2: Different VLM backbone (InternVL3.5-1B)}}                                                                                 \\
                6M\textsubscript{base}                        & -      & 85.76                  & 89.91                  & 81.22                  & 85.31                  \\
                6M                                            & \cmark & 89.81                  & 92.99                  & 84.76                  & 90.21                  \\
                2M                                            & \cmark & 88.62                  & 91.08                  & 82.91                  & 88.84                  \\
                1M                                            & \cmark & 84.93                  & 88.81                  & 78.25                  & 83.29                  \\
                Base                                          & -      & 34.81                  & 56.72                  & 29.24                  & 51.64                  \\
                \midrule
                \multicolumn{6}{l}{\textit{Exp3: Necessity of curriculum training (HunyuanOCR)}}                                                                           \\
                6M                                            & \xmark & 89.54$_{\down{-0.57}}$ & 92.41$_{\down{-0.56}}$ & 84.12$_{\down{-0.47}}$ & 89.40$_{\down{-0.62}}$ \\
                2M                                            & \xmark & 87.41$_{\down{-2.33}}$ & 90.33$_{\down{-2.55}}$ & 81.69$_{\down{-2.68}}$ & 87.31$_{\down{-2.69}}$ \\
                1M                                            & \xmark & 83.24$_{\down{-3.97}}$ & 86.91$_{\down{-3.56}}$ & 78.82$_{\down{-3.30}}$ & 83.52$_{\down{-4.21}}$ \\
                \bottomrule
            \end{tabular}
        }
    }
    \vspace{-3mm}
    \label{tab:further_experiments}
\end{table}

%% file: table/hard_sample_analysis.tex
\begin{table}[t!]
    \centering
    \captionsetup{font={small}}
    \caption{
        \textbf{Fine-grained analysis on hard samples of TableVerse-5K.}
        We isolate the 1{,}021 ``hard'' samples whose TEDS-S is below 0.75 after SFT. These samples are structurally more complex (avg. 19.31 rows / 9.52 cols) than the full set (14.39 rows / 7.13 cols). Compared with the vanilla RL baseline, our reward design yields substantially larger gains on these structurally challenging cases.
    }
    \vspace{-0.6em}
    {\renewcommand{\arraystretch}{0.95}
        \setlength{\tabcolsep}{10pt}
        \begin{tabular}{lcc}
            \toprule
            \textbf{Setting}         & \textbf{TEDS}  & \textbf{TEDS-S} \\
            \midrule
            StrucTab-SFT             & 0.415          & 0.514           \\
            Vanilla RL (TEDS-S)      & 0.524          & 0.623           \\
            \midrule
            \textbf{StrucTab (Ours)} & \textbf{0.572} & \textbf{0.682}  \\
            \bottomrule
        \end{tabular}
    }
    \vspace{-3mm}
    \label{tab:hard_sample_analysis}
\end{table}

%% file: section/8_appendix/8_6_discussions.tex
\section{Discussions}
\label{sec:supp_discussions}

In this section, we provide further analysis of several design choices behind the proposed StrucTab framework. While the main paper demonstrates the overall effectiveness of StrucTab, certain behaviors and mechanisms merit deeper examination. We focus on three key aspects: (a) the verification experiment that motivates the 1D Probe structural reward, (b) the motivation and feasibility of our Anchor-Guided Destylization strategy, and (c) a comparative analysis of training and inference efficiency.

\subsection{Early Error Correction Verification}
\label{subsec:supp_early_error}

We detail the verification experiment that motivates the 1D Probe structural reward, whose results are summarized in~\cref{subfig:error_rate_and_logp_comparison}. The goal is to test a simple hypothesis: due to the autoregressive nature of VLMs, an early structural mistake tends to propagate and amplify into later errors, so correcting the earliest structural deviation should substantially improve the subsequent generation.

\mypara{Probe Procedure.}
For each benchmark sample, we first let the model parse the table and locate the first structural error in its prediction. Following the same linearization as the 1D Probe, both the prediction and the reference are stripped of cell contents and flattened into row-major cell sequences; the first position where the two sequences disagree in row index, tag name, or cell attributes is taken as the first structural error. We then construct a corrected prefix by replacing the erroneous cell with its reference counterpart, feed this prefix back to the model as a teacher-forced assistant context, and let the model continue generating the remainder of the table. To isolate the effect of the single correction, decoding uses greedy sampling while recording per-token log-probabilities.

\mypara{Metrics.}
In the region after the correction point, we compare two quantities between the original and the corrected generations: (i) the \emph{subsequent structural error rate}, defined as the fraction of cells after the correction point whose structure mismatches the reference, and (ii) the \emph{generation confidence}, measured as the average token log-probability over the same region. As reported in~\cref{subfig:error_rate_and_logp_comparison}, correcting only the first structural error reduces the subsequent structural error rate by 75.25\% and increases the generation confidence by 27.85\% in log-probability. This confirms that early structural errors are the dominant source of downstream degradation, directly motivating a reward that explicitly rewards the longest correct structural prefix, namely the 1D Probe.

\subsection{Anchor-Guided Destylization}
\label{subsec:supp_destylization}

In this section, we elaborate on the motivation of Anchor-Guided Destylization.

\mypara{Stable RL Requires Clear and Consistent Training Signals.}
During RL training, we use high-quality real-world table images collected from multiple sources. Although these datasets are carefully curated, annotation inconsistencies inevitably arise due to different human labeling conventions. As a result, visually identical elements may appear in multiple syntactic forms, especially for mathematical expressions. Such stylistic variations introduce ambiguity into reward computation and can destabilize the reinforcement learning process.

\mypara{OCR Models Excel at Parsing Standardized Table Images.}
Real-world table images often contain diverse visual styles, including varying borders, fonts, and layouts, which increases parsing difficulty. In contrast, synthetically rendered tables (e.g., images generated from HTML via a browser engine) typically follow consistent font and layout conventions, making them easier to parse. Moreover, rendered document images are extensively used during the pretraining of modern OCR models to provide foundational perception and parsing capabilities~\cite{TRivia_2025, HunyuanOCR_2025, FD_RL_2026, MonkeyOCR_v1_5_2025}. As a result, existing OCR models are particularly effective at parsing standardized table images.

To verify this observation, we conducted a preliminary experiment. We first render tables from the open-source benchmarks using the rendering pipeline described in~\cref{alg:render_HTML_to_PNG}. The rendered images are then parsed by the open-source HunyuanOCR model. As shown in~\cref{tab:render_ocr_comparison}, parsing performance improves significantly on rendered images across multiple benchmarks, confirming the advantage of standardized visual inputs for OCR-based table parsing.

\begin{table}[h]
    \centering
    \small
    \vspace{-1.0em}
    \caption{Comparison of performance on original images and rendered images.}
    \vspace{-0.4em}
    \label{tab:render_ocr_comparison}
    {\renewcommand{\arraystretch}{0.95}
        \setlength{\tabcolsep}{6pt}
        \resizebox{\columnwidth}{!}{
            \begin{tabular}{lcccccccc}
                \toprule
                \multirow{2}{*}[-0.4mm]{Setting} &
                \multicolumn{2}{c}{OmniDocBench} &
                \multicolumn{2}{c}{CC-OCR}       &
                \multicolumn{2}{c}{OCRBench v2}  &
                \multicolumn{2}{c}{Average}
                \\
                \cmidrule(lr){2-3}
                \cmidrule(lr){4-5}
                \cmidrule(lr){6-7}
                \cmidrule(lr){8-9}
                                                 & TEDS  & TEDS-S & TEDS  & TEDS-S & TEDS  & TEDS-S & TEDS  & TEDS-S \\
                \midrule
                Original images                  & 95.74 & 97.71  & 81.44 & 86.91  & 82.34 & 87.74  & 86.69 & 90.95  \\
                Rendered images                  & 97.38 & 98.99  & 89.22 & 96.95  & 85.89 & 93.03  & 90.44 & 95.83  \\
                \bottomrule
            \end{tabular}
        }
    }
    \vspace{-1.0em}
\end{table}

\mypara{Anchor-Guided Destylization.}
Based on this observation, we render both sampled predictions and reference HTML using a unified rendering pipeline to obtain standardized table images. These images are then parsed by the anchor OCR model, aligning all outputs to a consistent structural representation. TEDS scores are subsequently computed on the normalized parsing results, ensuring that the reward focuses on structural and semantic correctness rather than superficial stylistic differences. To make the anchor parsing fully deterministic, we decode the anchor OCR model with greedy sampling (temperature $0$, top-$p$ $0.95$, top-$k$ $64$, repetition penalty $1.0$) under a fixed random seed. Combined with the standardized rendering pipeline, this determinism allows the de-stylized references to be pre-computed offline and reused throughout RL training, avoiding repeated anchor parsing of the reference side.

\subsection{Comparative Analysis of RL Efficiency}
\label{subsec:supp_RL_efficiency}

As shown in~\cref{tab:Uni_TabRL} of the main paper, RL-based training consistently outperforms StrucTab-SFT, although computational cost varies across reward configurations. Rule-based methods (a) and (d) introduce negligible overhead, while configuration (d) with our 1D Probe structural reward achieves substantial performance gains. In contrast, setting (b), which uses a VLM judge (Qwen2.5-VL-72B), increases training time by 14.7\%. Our Anchor-Guided Destylization (c) provides a favorable trade-off between efficiency and effectiveness. By deploying the lightweight HunyuanOCR model with vLLM~\cite{vLLM_2023} and EAGLE-3~\cite{Eagle_3_2025} speculative decoding, we significantly reduce destylization latency, increasing total training time by only 7.4\%, roughly half the cost of the VLM-based judge. This result demonstrates that our decomposed reward framework achieves both strong performance and high computational efficiency.

\subsection{Inference Efficiency of Sequential Reasoning}
\label{subsec:inference_efficiency}

In terms of inference efficiency, the structured reasoning paradigm introduces only marginal overhead. Compared with direct image-to-HTML generation, the sequential reasoning stage increases the average HTML parsing length by merely 7.3\% (from 862 to 925 average tokens), which is negligible relative to the substantial accuracy gains. Since the backbone remains a compact 1B-parameter VLM, the overall inference efficiency stays high, and the additional reasoning tokens do not pose a practical bottleneck for deployment.